\documentclass[journal]{IEEEtran}

\ifCLASSINFOpdf
\else
\fi

\usepackage{todonotes}

\usepackage{amsmath,amsfonts,epsfig,graphicx}
\usepackage{subfig}
\usepackage{color,soul}
\usepackage{epstopdf}
\usepackage{tikz}
\usepackage{balance}
\usepackage{url}
\usepackage{multirow,hhline}
\usepackage{placeins}
\usepackage{wasysym}
\usepackage{amssymb}
\usepackage{pifont}

\usepackage[nomessages]{fp}

\usepackage{pgfplots}
\pgfplotsset{compat=newest}
\usetikzlibrary{plotmarks}
\usetikzlibrary{matrix}

\newcommand{\ru}{\rule{0mm}{3mm}}

\newcommand{\DL}{D_{\lambda}^{(\rm K)}}
\newcommand{\DS}{D_{\rm S}}
\newcommand{\DR}{D_{\rho}}

\begin{document}

\title{Pansharpening by convolutional neural networks in the full resolution framework}

\author{
Matteo~Ciotola, Sergio~Vitale, Antonio~Mazza, Giovanni~Poggi, Giuseppe~Scarpa, 
\thanks{
M.Ciotola, A.Mazza, G.Poggi, and G.Scarpa are with the Department of Electrical Engineering and Information Technology, University Federico II, Naples, Italy, e-mail: \{firstname.lastname\}@unina.it.
S.Vitale is with the Department of Science and Technology, University Parthenope, Naples, Italy, e-mail: sergio.vitale@uniparthenope.it.}
}
\maketitle

\begin{abstract}
In recent years, there has been a growing interest in deep learning-based pansharpening.
Thus far, research has mainly focused on architectures.
Nonetheless, model training is an equally important issue.
A first problem is the absence of ground truths, unavoidable in pansharpening.
This is often addressed by training networks in a reduced resolution domain and using the original data as ground truth, relying on an implicit scale invariance assumption.
However, on full resolution images results are often disappointing, suggesting such invariance not to hold.
A further problem is the scarcity of training data,
which causes a limited generalization ability and a poor performance on off-training test images.

In this paper, we propose a full-resolution training framework for deep learning-based pansharpening.
The framework is fully general and can be used for any deep learning-based pansharpening model.
Training takes place in the high-resolution domain, relying only on the original data, thus avoiding any loss of information.
To ensure spectral and spatial fidelity, a suitable two-component loss is defined.
The spectral component enforces consistency between the pansharpened output and the low-resolution multispectral input.
The spatial component, computed at high-resolution, maximizes the local correlation between each pansharpened band and the panchromatic input.
At testing time, the target-adaptive operating modality is adopted, achieving good generalization with a limited computational overhead.

Experiments carried out on WorldView-3, WorldView-2, and GeoEye-1 images show that methods trained with the proposed framework
guarantee a pretty good performance in terms of both full-resolution numerical indexes and visual quality.
\end{abstract}

\IEEEpeerreviewmaketitle

\section{Introduction}
Given the ever-increasing number of satellites acquiring images of the Earth,
data fusion is becoming a key asset in remote sensing,
enabling cross-sensor \cite{Gargiulo2018, Errico2014}, cross-resolution \cite{Vivone2015} or cross-temporal \cite{Gaetano2014} analysis and information extraction.
Due to technological constraints, many Earth observation systems, such as GeoEye, Plaiades or WorldView,
acquire a single full resolution panchromatic band (PAN), responsible for the preservation of geometric information,
along with a multispectral image (MS) at lower spatial resolution, with rich spectral information.
A multi-resolution fusion process, called pansharpening, is then employed
to estimate a full resolution multispectral image from the original PAN and MS components \cite{Aiazzi2006, Vivone2015}.

Pansharpening is a challenging task, object of intense research for three decades but still far from being solved,
also because of the continuously increasing resolutions at which new generation satellites operate.
Several approaches and a large number of methods have been proposed over the years.

In the component substitution (CS) approach \cite{Shettigara1992},
the multispectral image is transformed in a suitable domain, one of its components is replaced by the spatially rich PAN, and the image is transformed back in the original domain.
If only three bands are concerned, the Intensity-Hue-Saturation (IHS) transform can be used,
with the intensity component replaced by the panchromatic band \cite{Tu2001}.
The method is generalized in \cite{Tu2004} (GIHS) to handle a larger number of bands.
Many other transforms have been considered for CS,
including principal component analysis \cite{Chavez1989}, Brovey transform \cite{Gillespie1987} and Gram-Schmidt (GS) decomposition \cite{Laben2000}.
More recently, adaptive CS methods have also been introduced, such as
the advanced versions of GIHS and GS \cite{Aiazzi2007},
the partial replacement CS method (PRACS) \cite{Choi2011}, or
the band-dependent spatial detail (BDSD) injection method and its variants \cite{Garzelli2008,Garzelli2015, Vivone2019}.

With the multiresolution analysis (MRA) approach \cite{Ranchin2000}, instead, pansharpening is addressed from the spatial perspective.
These methods extract the high frequency spatial details through a multi-resolution decomposition, such as
decimated or undecimated wavelet transforms \cite{Nunez1999, Ranchin2000, Otazu2005, Khan2008},
Laplacian pyramids \cite{Aiazzi2002, Aiazzi2003, Aiazzi2006, Lee2010, Restaino2017}, or
other nonseparable transforms, {\it e.g.}, contourlet \cite{Shah2008}.
Extracted details are then properly injected into the resized MS component.

A further set of methods address the pansharpening problem through the variational optimization (VO) of suitable acquisition or representation models.
In \cite{Vivone2015a}, the optimization functional involves the degradation filters mapping high-resolution to low-resolution images,
whereas \cite{Vicinanza2015} focuses on the sparse representations of injected details.
Palsson {\it et al.} proposed several methods of this class,
using a total variation regularized least square formulation in \cite{Palsson2014},
defining a maximum {\em a posteriori} problem in \cite{Palsson2015} and,
very recently \cite{Palsson2020}, looking for low-rank representations of the joint PAN-MS pair organized in a suitable matrix.
Other methods do not fit the above categories and can be roughly classified as
statistical~\cite{Fasbender2008, Zhang2012, Meng2015, Shen2016, Zhong2017},
dictionary-based \cite{Li2011, Li2013, Zhu2013, Cheng2014, Zhu2016, Hong2019},
or matrix factorization approaches \cite{Yokoya2012, Lanaras2015, Hong2019b}.
The reader is referred to \cite{Vivone2015} for a more comprehensive review.

In recent years, a paradigm shift from model-based to data-driven approaches has revolutionized all fields of image processing,
from computer vision \cite{Krizhevsky2012, Dong2016, He2017, Lateef2019, Zhao2019} to remote sensing \cite{Yang2017, Scarpa2018, Benedetti2018, Mazza2019}.
In pansharpening
the first method based on convolutional neural networks (CNN) was proposed by Masi {\it et al.} in 2016 \cite{Masi2016},
after which many more followed in a few years' span \cite{Yang2017, Wei2017a, Wei2017L, Rao2017, Masi2017, Azarang2017, Yuan2018, Liu2018, Shao2018, Vitale2018, Zhan2019,Dong2021A,Dong2021B}.
It seems safe to say that deep learning is currently the most popular approach for pansharpening.
Nonetheless, it suffers from a major problem: the lack of ground truth data for supervised training.
In fact, multi-resolution sensors can only provide the original MS-PAN data, downgraded in space or spectrum,
never their high-resolution versions, which remain to be estimated.

A widespread solution to this problem is to perform a resolution shift.
The PAN-MS data undergo a downsampling process, after which they are used as input samples to train a network where the original MS data play the role of ground truth.
By doing so, the network is trained in a fully supervised manner, although in a lower-resolution domain.
Eventually, it will be used for pansharpening the original data.
Therefore, this solution relies on a sort of scale-invariance assumption: a network trained at low resolution is expected to work equally well at high resolution.
That this hypothesis holds up, however, is by no means obvious.

In the literature, this problem is well known \cite{Wald97} and, in fact,
great attention is devoted to mimic the sensor modulation transfer functions (MTF) to ensure a correct downgrading of data.
Even with an ideal scaling process, however, an inherent information gap exists between scales.
For example,
objects whose typical size amounts to a few pixels at the original resolution will necessarily lose their shape when brought at low resolution.
There is no hope that a network trained at reduced resolution will ``experience'' such tiny geometries.
Not surprisingly,
networks trained with this approach work very well on reduced-resolution data,
but show significant quality losses on full-resolution target data \cite{Masi2016, Yang2017, Wei2017a, Scarpa2018a}.
Interestingly,
these problems have often been overlooked precisely because, in the absence of ground truth,
it is not possible to objectively measure performance at target resolution.

A further limit of deep learning-based pansharpening is the endemic scarcity of remote sensing training data.
Due to the high cost of multi-resolution data,
networks are usually trained on just a few images which, however large,
cannot ensure an adequate diversity in terms of geographical position, territorial conformation, atmospheric conditions,
acquisition geometry, direction and intensity of light, etc.
As a consequence, such networks will hardly generalize to images acquired by sensors not seen in training, or even just to different-looking images.

Motivated by these considerations,
in this paper we propose a new framework for training pansharpening models in the high resolution domain.
Networks are trained using the original PAN-MS pairs as input, at their native resolution, with no downgrading and hence no loss of information.
To obviate the lack of a ground truth, a new {\it ad hoc} loss is defined, which weights suitably defined indicators of spatial and spectral consistency.
These indicators are computed by comparing the pansharpened output with the original PAN and MS components in their respective domains.
In addition, to ensure correct operations on images with the most diverse characteristics,
notwithstanding the limited datasets available for training,
we use the target-adaptive modality proposed originally in \cite{Scarpa2018a} which fine-tunes the network on the fly to the target image.
Finally, it is worth underlining that the proposed learning framework is fully general and can be used for any deep learning-based pansharpening model.
Experiments with three state-of-the-art CNN-based pansharpening models on images acquired by different multi-resolution sensors,
demonstrate the broad and seamless applicability of this framework,
as well as the significant quality improvements ensured by high-resolution training.

In summary,
the main innovative contribution of this work is the proposal of a new fully unsupervised framework
which allows training deep learning-based pansharpening models at high-resolution.
To validate the proposal, we re-train several state-of-the-art methods in the new framework and carry out a wide range of experiments on images acquired by several sensors.
Moreover, to ensure research reproducibility,
we publish online a user-friendly software package for high-resolution training and testing of pansharpening networks,
together with several trained models.\footnote{GitHub repository: \url{https://github.com/matciotola/Z-PNN}}

Following this Introduction,
in Section II we account for related work,
in Section III describe the proposed full resolution training framework,
in Section IV, present experimental result,
and eventually in Section V draw conclusions.

\newcommand{\DD}{{\cal D}}
\newcommand{\BB}{{\cal B}}
\newcommand{\GG}{{\cal G}}
\newcommand{\LL}{{\cal L}}
\newcommand{\wM}{{\widehat{M}}}

\section{Related work}

In recent years there has been a growing awareness that the resolution-shift approach to training pansharpening networks has inherent weaknesses and may cause a performance cap.
Starting in 2020, several papers have begun to address this issue
and to propose new solutions that carry out training, at least partially, in the high resolution domain.

The first of these papers \cite{Vitale2020}, to the best of our knowledge, has been proposed by some of the authors of the present work.
Training is carried out in fully supervised modality with a loss that includes both reduced resolution and full resolution terms.
At low resolution, the resolution-shift approach is used, with the original MS acting as ground truth.
At high resolution, instead, the output of the MTF-GLP-HPM model-based algorithm \cite{Aiazzi2003} takes the role of ground truth.
Indeed, this algorithm is known to ensure a very good preservation of high resolution details,
which justifies using it as a proxy of the unknown ground truth for the only purpose of improving spatial quality.
Needless to say, spatial accuracy cannot be better than that of the auxiliary algorithm, certainly non optimal.
An enhanced version of the method was later proposed in \cite{Vitale2020a},
with a spatial loss terms relying also on the preservation of spatial gradients.
Eventually, both versions provide only minor improvements with respect to methods relying on reduced-resolution learning schemes.
A further development \cite{Ciotola2021} concerns the fusion of high and low-resolution spectral bands in Sentinel-2 images, a closely related task.

In \cite{Luo2020} a rather complex residual network is proposed, trained at high resolution.
Features extracted from the PAN are used in a sequence of fusion units to refine the high-pass details extracted from the upsampled MS.
The loss includes spatial and spectral terms, compounding both Euclidean norm and structural similarity (SSIM),
together with a term depending on a no-reference quality index.
Despite the stated goal of overcoming the resolution-shift approach,
these loss terms depend heavily on cross-scale consistency indexes, thereby reintroducing a sort of scale invariance assumption.
In addition, an MS-to-PAN operator is used (called it $\GG$, in Section III)
which combines linearly the MS bands through coefficients estimated, again, at low resolution.
Experimental results seem promising, but training and test data come from the same scene and do not allow to test generalization ability.

A deep CNN, called UPSNet, comprising 28 residual blocks plus two adaptation blocks, is proposed in \cite{Seo2020}.
Loss terms are computed exclusively on high resolution data, with spatial accuracy pursued by working on the PAN gradients.
However, they depend again on some ill-defined pieces of information, such as "grayed" or upsampled versions of the MS.
To make up for errors originated by such grayed MS, a further loss is introduced which, however, involves also non-differentiable functions.
Despite these shortcomings, good quality pansharpened images are obtained, although a bit oversmoothed.

A group of recent papers on this topic rely on generative adversarial networks (GAN).
Indeed, GANs seem to fit very well the pansharpening task.
The generator may be charged with the task of producing the high resolution output starting from the available PAN and MS,
while two dedicated discriminators validate the quality of results by
comparing the panchromatic and low-resolution projections of the output with the original counterparts.
None of these processes require a resolution shift.
PanGAN \cite{Ma2020}, PercepPAN \cite{Zhou2020}, and PGMAN \cite{Zhou2021}, all follows this approach, with minor variations.
However, despite the elegant formulation, results turn out to be much below expectations,
with visible spectral aberrations (PanGAN, PercepPAN) or spatial blurring (PGMAN).
Arguably, such poor results may be due to seemingly minor inaccuracies that disrupt the delicate training process of GANs.
Such inaccuracies include
the use of arbitrary MS-to-PAN linear projections with coefficients estimated on unrepresentative data
and
imperfect MS interpolation.

Despite their obvious value, these contributions present some common limits and flaws:
\begin{itemize}
\item   they concern individual pansharpening methods trained at high resolution, not a general training framework;
\item   rely heavily on potentially detrimental cross-scale processing steps, such as arbitrary forms of interpolation or decimation, or MS-to-PAN conversions;
\item   generalize poorly to images with characteristic not seen in training, especially if acquired by sensors not represented in the dataset;
\item   methods and results are hardly reproducible due to the lack of software code online.
\end{itemize}

On the contrary, we propose a high-resolution training {\em framework},
applicable to any deep learning-based network, even if designed originally for reduced resolution training.
We minimize cross-scale processing, limited to a single downsizing step for loss computation.
Correct operations on the most diverse images is ensured by the target-adaptive modality.
Finally, we make all our software available online to allow easy reproduction of results and easy development of further improvements.

\section{Proposed full-resolution training framework}
\label{sec:proposed}

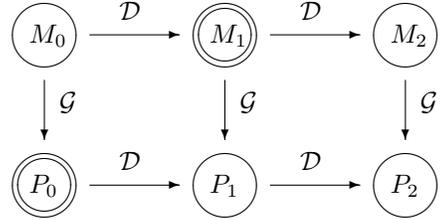
\begin{figure}
\centering
\newcommand{\DDx}{{\cal D}}
\newcommand{\GGx}{{\cal G}}
\newcommand{\LLx}{{\cal L}}
\newcommand{\wMx}{{\widehat{M}}}
\setlength{\unitlength}{1mm}
\begin{picture}(88,40)(000,000)
\put(16,30){\circle{8}}
\put(40,30){\circle{8}} \put(40,30){\circle{7.5}}
\put(64,30){\circle{8}}
\put(16,10){\circle{8}} \put(16,10){\circle{7}}
\put(40,10){\circle{8}}
\put(64,10){\circle{8}}
\put(14,29){$M_0$}
\put(38,29){$M_1$}
\put(62,29){$M_2$}
\put(14,09){$P_0$}
\put(38,09){$P_1$}
\put(62,09){$P_2$}
\put(22,30){\vector(1,0){12}}
\put(26,32){$\DDx$}
\put(46,30){\vector(1,0){12}}
\put(50,32){$\DDx$}
\put(22,10){\vector(1,0){12}}
\put(26,12){$\DDx$}
\put(46,10){\vector(1,0){12}}
\put(50,12){$\DDx$}
\put(16,24){\vector(0,-1){8}}
\put(18,20){$\GGx$}
\put(40,24){\vector(0,-1){8}}
\put(42,20){$\GGx$}
\put(64,24){\vector(0,-1){8}}
\put(66,20){$\GGx$}
\end{picture}
\caption{Images and scales involved in the pansharpening process.
The only available pieces of information are the full-resolution panchromatic image, $P_0$, and the low-resolution multispectral image, $M_1$,
from which the target high-resolution multispectral image, $M_0$, is estimated.
Deterministic (only partially known) operators, $\DD$ and $\GG$, relate images with their spatially or spectrally downgraded versions.}
\label{fig:scales}
\end{figure}

In the following, we will use $M$ and $P$, respectively, to denote multispectral and panchromatic images.
A subscript will indicate their spatial scale, with 0 associated with the highest resolution, and a fixed resolution ratio $R$ between scales $n$ and $n+1$, for each $n$.
The relationship between these images is depicted in Fig.~\ref{fig:scales}
where it is also assumed that
low resolution images can be obtained from their higher resolution versions through a deterministic operator, $\DD$, and
panchromatic images from the corresponding multispectral ones through another operator, $\GG$.
This assumption holds with good approximation for the downscaling operator, $\DD$,
while MS-to-PAN operators, though often used in applications, are necessarily far from ideal because of the sensors' physical characteristics.
Of course such operators imply a loss of information and hence are not invertible.

In multi-resolution remote sensing, $M_1$ and $P_0$ are the only available pieces of information (the MS-PAN pair)
and in fact the goal of pansharpening is to estimate the unknown high-resolution multispectral image $M_0$ from these spatially and spectrally degraded images,
\begin{equation}
    \wM_0 = \phi_0(M_1,P_0)
\end{equation}
In {\em deep learning-based} pansharpening, the estimator $\phi_0$ is learned from a suitable collection of training data.
This would be a standard task if complete training data were available, that is, for each training input pair $(M_1^i,P_0^i)$, the corresponding desired output $M_0^i$ was also provided.
However, this is not the case, no full-resolution multispectral images are available to be used as ground truth.

\begin{figure*}
\centering
\tikzstyle{Circle}=[circle, draw, fill=white, minimum size = 22pt, inner sep=0pt]
\tikzstyle{DoubleCircle}=[circle, double, double distance=1pt, draw, fill=white,minimum size = 20.5pt, inner sep=0pt]
\tikzstyle{Rectangle}=[rectangle,draw, minimum width=20pt, minimum height=20pt,inner sep=0pt]
\tikzstyle{Concat}=[circle, draw, minimum size=10pt, inner sep=0pt]
\tikzstyle{arr}=[ -stealth, shorten <=3pt, shorten >=3pt]
\tikzstyle{CNN}=[rectangle, very thick,  draw, minimum width=45pt, minimum height=25pt,inner sep=0pt]
\newcommand{\D}{$\mathcal{D}$}
\newcommand{\step}{2}
\begin{tabular}{ccc}
\begin{tikzpicture}[node distance=2cm]
\node[DoubleCircle](P0) at (0,0) {$P_0$};
\node[DoubleCircle](M1) at (0.5*\step,0) {$M_1$};
\node[Rectangle](L)     at (2.5*\step,0) {$\mathcal{L}$};
\node[Circle](M2)       at (0.5*\step,-0.8*\step) {$M_2$};
\node[Circle](P1)       at (0,-0.8*\step) {$P_1$};
\node[Concat](c)        at (0.5*\step,-1.5*\step) {\bf c};
\node[CNN] (cnn)        at (1.5*\step, -1.5*\step) {\parbox{40pt}{\centering \small Network \normalsize $\phi_1$}};
\node[Circle](HM1)      at (2.5*\step,-1.5*\step) {$\widehat{M}_1$};
\draw[arr] (P0)--node[right]{\D}(P1);
\draw[arr] (M1)--node[right]{\D}(M2);
\draw[arr] (M2)--(c);
\draw[arr] (P1)|-(c);
\draw[arr] (c)--(cnn);
\draw[arr] (cnn)--(HM1);
\draw[arr] (M1)--(L);
\draw[arr] (HM1)--(L);
\end{tikzpicture}
& \hspace{12mm} & 
\begin{tikzpicture}[node distance=2cm]
\node[DoubleCircle](P0) at (0,0) {$P_0$};
\node[Rectangle](LS) at (3*\step,0) {$\mathcal{L}_{\rm S}$};
\node[DoubleCircle](M1) at (0.5*\step,-0.3*\step) {$M_1$};
\node[Rectangle](LL) at (2.5*\step,-0.3*\step) {$\mathcal{L}_\lambda$};
\node[Concat](c) at (0.5*\step,-1.5*\step) {\bf c};
\node[CNN] (cnn) at (1.5*\step, -1.5*\step) {\parbox{40pt}{\centering \small Network \normalsize $\phi_0$}};
\node[Circle](HM0) at (2.5*\step,-1.5*\step) {$\widehat{M}_0$};
\draw[arr] (P0)|-(c);
\draw[arr] (M1)--(c);
\draw[arr] (c)--(cnn);
\draw[arr] (cnn)--(HM0);
\draw[arr] (M1)--(LL);
\draw[arr] (P0)--(LS);
\draw[arr] (HM0)--node[right]{\D}(LL);
\draw[arr] (HM0)-|(LS);
\end{tikzpicture}
\\
\end{tabular}
\caption{Wald-like (left) and proposed (right) training frameworks.
In the first case, training takes place in the reduced resolution domain, MS and PAN are immediately downgraded, and the latter is not used to compute the loss.
In the proposed framework, only the original high-resolution PAN and MS are used for training, and they are both used to compute the loss.
}
\label{fig:wald}
\end{figure*}
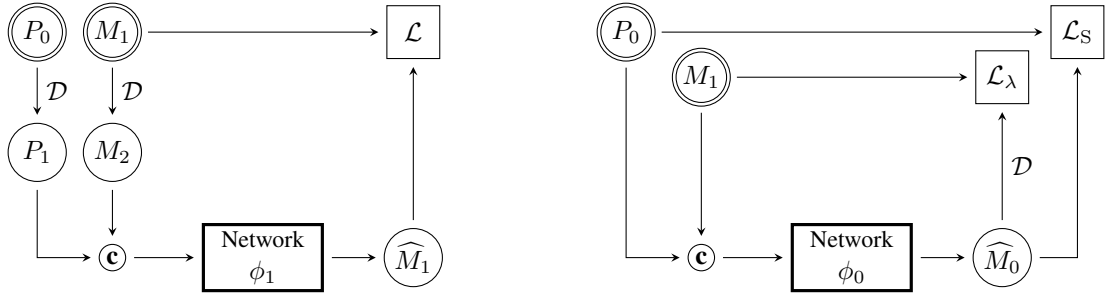

Most deep learning-based pansharpening methods proposed thus far \cite{Masi2016, Wei2017L, Yang2017, Scarpa2018a} have circumvented this problem
by means of a domain shift approach, known as Wald's protocol \cite{Wald97}, which allows to assess their synthesis ability.
All available images in the dataset are downscaled to the next lower resolution level
\begin{equation}
    M_2^i =  \DD(M_1^i), \hspace{6mm} P_1^i = \DD(P_0^i)
\end{equation}
For these pairs, the original multispectral images, $M_1^i$, represent the perfectly known ground truth.
Therefore, a conventional training procedure can be used to estimate the network weights, that is, the pansharpening function $\phi_1$.
This function is eventually used to perform pansharpening at the original scale.
A block diagram of this training procedure is shown in the left part of Fig.~\ref{fig:wald}.

Of course, underlying this approach is the assumption that the same network can operate equally well at low resolution and high resolution, that is $\phi_1 \simeq \phi_0$.
This is a convenient approximation, but experimental evidence accumulated over the years prove it to be largely inaccurate.
Networks trained under Wald's resolution downgrading protocol work very well on the low resolution images they have been trained for, but only fairly well \cite{Vivone2020} on the full resolution images.
In practice, there is a significant domain mismatch between low-resolution and high-resolution pansharpening.

Before proposing our alternative training framework, let us justify intuitively the unsatisfactory behavior of the resolution shift solution.
The fundamental observation is that the network, during the entire training process, never sees the full resolution data.
In particular, the panchromatic images, the only data available at full resolution, are immediately resized, causing an irrecoverable loss of information.
To fully realize the importance of this loss, one should also keep in mind that these images are acquired at a fixed resolution.
For example, all panchromatic images provided by the WorldView-3 sensor have a spatial resolution of 0.31m.
At this resolution, a number of small urban objects, like cars, traffic signs, and so on, are fully characterized with well-defined geometric shapes.
With the help of low-resolution spectral information, they can be accurately recovered.
However, with the resolution shift approach,
the network sees only images of much lower resolution, 1.24m (with 4.96m multispectral) where these tiny objects loose completely their shape, reducing to a very few pixels or even sub-pixels.
Contrary to what happens in super-resolution,
there is no 8cm-resolution WorldView-3 image available to make up for this loss of information.

An additional problem is that resized images are much smaller than the original ones, providing much less data for training.
Sticking to the WorldView-3 example, at low resolution there are 16 times less pixels than at full resolution.
Considering the scarcity of training data, due to the restrictive policies of most data providers, this turns out to be a non-negligible drawback.

These considerations, together with experimental results much below expectations, motivate our proposal of a full-resolution training framework.
We will train pansharpening networks using the original data, thereby including full-resolution panchromatic images.
Clearly, we must do without the ground truth images, which do not exist.
Therefore, the cornerstone of our proposal is the definition of a new loss function that takes the role of the conventional full-reference loss.

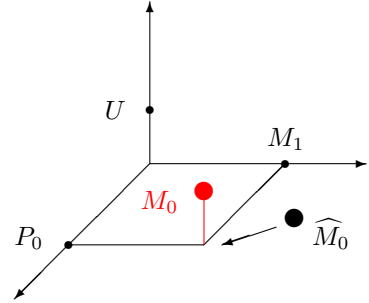
\begin{figure}
\centering
\setlength{\unitlength}{1.2mm}
\begin{picture}(42,37)(000,000)
\put(16,18){\vector(+1,+0){24}}
\put(16,18){\vector(-1,-1){15}}
\put(16,18){\vector(+0,+1){18}}
\put(7,09){\circle*{1}} \put(1,09){$P_0$}
\put(31,18){\circle*{1}} \put(29,20){$M_1$}
\put(16,24){\circle*{1}} \put(11,23){$U$}
\put(32,12){\circle*{2}} \put(34,09){$\widehat{M}_0$}
\put(30,11){\vector(-3,-1){6}}
\put(22,09){\line(-1,+0){15}}
\put(22,09){\line(+1,+1){09}}
{\color{red}
\put(22,15){\circle*{2}} \put(15,13){$M_0$}
\put(22,09){\line(+0,+1){06}} 
}
\end{picture}
\caption{Graphical sketch of the ideal proposed approach.
Lacking a ground truth, the pansharpening process aims at generating an image, $\wM_0$, whose projections coincide with the known original data, $P_0$ and $M_1$.
An unknown residual, $U$, orthogonal to this plane, remains unaccounted for.
Our conjecture is that this latter component is small.}
\label{fig:xyz}
\end{figure}

Since we lack the full-resolution reference, $M_0$, we use the next most valuable pieces of information, that is, its projections on the low-resolution and panchromatic domains, $M_1$ and $P_0$.
The network output $\wM_0$ is compared with these two references, in their respective domains, to ensure spectral and spatial consistency.
Accordingly, the proposed loss becomes
\begin{eqnarray}
    \LL(M_1,P_0;\wM_0) = \hspace{40mm} \nonumber & \\
            \LL_{\lambda}(M_1;\DD(\wM_0)) + \beta \LL_S(P_0;\GG(\wM_0)) &
\label{eq:theoretical_loss}
\end{eqnarray}
with $\beta$ a suitable parameter that weighs the spectral and spatial loss terms.

Fig.~\ref{fig:xyz} illustrates our approach geometrically.
The target image $M_0$ (red dot) is regarded as the combination of its $M_1$ and $P_0$ projections
plus a third unknown component (call it $U$) which cannot be explained by neither of the former two.
By minimizing the loss of Eq.(\ref{eq:theoretical_loss}) we are pushing the estimate $\wM_0$ (black dot) towards the projections of $M_0$ on the $(M_1,P_0)$ plane.
The origin of the third component has been critically explored in \cite{Thomas2008}, comparing alternative perspectives and assumptions.
Our working hypothesis is that this unpredictable part is indeed small and, therefore, our final estimate will be very close to the actual image.
It is left to the experimental results to say the final word in favor or against this hypothesis.
At the very least, with our approach we are not discarding any relevant data in the training process.

In the practical implementation, we depart slightly from the elegant symmetric formulation of Eq.(\ref{eq:theoretical_loss}).
Indeed, while the $\DD$ operator can be reasonably assumed to be known, such that $M_1=\DD(M_0)$,
there is no consensus in the literature on the exact form and even on the conceptual correctness of the $\GG$ operator.
To circumvent this problem, this operator is by-passed, here,
and the spatial loss term is computed as the sum of $B$ individual contributions, one for each spectral band of $\widehat{M}_0$.
Synthetically, the proposed loss reads as
\begin{eqnarray}
    \LL(M_1,P_0;\wM_0) = \hspace{40mm} \nonumber & \\
            \LL_{\lambda}(M_1;\DD(\wM_0)) + \beta \LL_S(P_0;\wM_0) &
\label{eq:actual_loss}
\end{eqnarray}

A block diagram of the proposed training procedure is shown in Fig.\ref{fig:wald}, next to the Wald-like training scheme with resolution downgrading, for easy comparison.
Visual inspection provides an immediate appreciation of the fundamental changes:
\begin{enumerate}
\item   in the Wald-like framework, $P_0$ is immediately downgraded and never used further, therefore high-resolution information is lost forever;
\item   in the proposed framework, instead, an additional spatial loss term $\LL_{\rm S}$ is introduced to take advantage of the information conveyed by the PAN;
\item   in the proposed framework, the only resolution downgrade takes place {\em after} pansharpening, and only for the purpose of comparison with the original MS.
\end{enumerate}
In the following two subsections, we describe in detail the spectral and spatial loss terms.

\subsection{Spatial loss}

The role of the spatial loss is to inject in the pansharpened image the high-resolution structures observed in the PAN.
Accordingly, the PAN can be used to perform a prediction, necessarily imperfect, of the output image bands,
and preferably a linear prediction, lacking any reasons to prefer more complex solutions.
Following this point of view,
here, we define the spatial loss term as a function of the correlation coefficient between the PAN ad the spectral bands of the output image.

Let $X$ and $Y$ be two equal-size single-band images, and let $\sigma^2_X$, $\sigma^2_Y$ and $\sigma_{XY}$ indicate their sample variances and covariance.
Then, the correlation coefficient between $X$ and $Y$ is defined as
\begin{equation}
    \rho_{XY} = \frac{\sigma_{XY}}{\sigma_X\sigma_Y}, \hspace{6mm} -1 \leq \rho_{XY} \leq 1
    \label{eq:correlation}
\end{equation}
The correlation coefficient indicates to what extent one image can be linearly predicted from the other,
with $|\rho|=1$ implying perfect predictability and $\rho=0$ total incorrelation.

Now, we expect to find in the pansharpened bands mostly the same spatial layout of the PAN, and therefore a strong correlation with it.
However, to preserve the spectral information, such a correlation cannot be unitary.
Actually, it can be expected to vary spatially and from band to band, as a function of the observed scene.
For example, in vegetated areas, we expect the PAN to have a strong correlation with the ``green'' band of the output, and a weaker correlation with other bands,
while the opposite will happen in other regions.
In rare cases, even negative correlations are observed, due to local contrast inversions between the PAN and some MS bands \cite{Thomas2008}.
This leads us to consider a three-dimensional spatial-spectral correlation field rather than a single coefficient.
So, in Eq.(\ref{eq:correlation}),
let $X$ be a square patch of size $\sigma\times\sigma$ extracted from the PAN at spatial location $(i,j)$ and
let $Y$ be the corresponding patch extracted from band $b$ of $\wM_0$,
then we obtain the three-dimensional correlation field
\begin{equation}
    \rho^\sigma(i,j,b) \; \triangleq \; \rho^\sigma_{P,\wM_0(b)}(i,j)
\end{equation}
which depends on spatial coordinates $(i,j)$, spectral coordinate $b$, and on the size parameter $\sigma$.

Now, we could think of defining a local spatial loss as
\begin{equation}
    \ell^\sigma(i,j,b) = 1-\rho^\sigma(i,j,b), \hspace{6mm} 0 \leq \ell \leq 2
\end{equation}
and the global spatial loss term as its average.
However, by doing so we would neglect the inherent spatial-spectral variability of the correlation mentioned above, and push it uniformly towards 1.
Therefore, to address this problem we define an auxiliary reference correlation field,
$\rho^{\sigma,{\rm ref}}(i,j,b)$,
computed between a low-pass filtered version of the PAN and an expanded version (plain interpolation) of the MS,
and define the local loss as
\begin{equation}
    \ell^\sigma(i,j,b) = \left\{ \begin{array}{ll}
            1-\rho^\sigma(i,j,b) & \rho^\sigma < \rho^{\sigma,{\rm ref}} \\
            0                    & \mbox{otherwise}
            \end{array} \right.
\end{equation}
The reference correlation field can be computed exactly from the available data and provides a rough approximation of the target correlation field.
A positive loss $\ell^\sigma=1-\rho^\sigma$ is incurred at site $(i,j,b)$ whenever the local correlation is too small, forcing the output band to follow the spatial layout of the PAN.
When $\rho^\sigma$ exceeds the reference value $\rho^{\sigma,{\rm ref}}$, however,
there is no further contribution to the global loss, and the network is free to optimize the output based on other inputs.

Although the use of correlation is certainly not new in pansharpening,
we point out that our approach is very different from what encountered in conventional methods.
Component substitution, for example, relies on the strong assumption of a perfect {\em global} correlation between the pansharpened MS bands and the PAN \cite{Thomas2008}.
When this assumption is violated, especially in the presence of occultation or contrast inversion phenomena,
strong spectral aberrations are observed.
In some traditional injection-based methods \cite{Ranchin2003}, instead, local correlation is used just to exert a consistency check.
Injection of PAN details takes place only when the local correlation is high, switching to a plain upsampling of the MS bands otherwise.
We assume a {\it generally} large local correlation between PAN and MS,
but verify our hypothesis on the reference field, $\rho^{\sigma,{\rm ref}}$, and leverage deep learning with a suitable loss to exploit this dependence.

\subsection{Spectral loss}

The spectral loss is computed in a straightforward manner
by comparing the low-resolution projection of the pansharpened image, $\DD(\wM_0)$, with its natural reference $M_1$,
\begin{equation}
    \LL_\lambda = \parallel \DD(\wM_0)-M_1\parallel_1
\end{equation}
where $\parallel\cdot\parallel_1$ indicates the $\ell_1$-norm.

As already said,
the low-resolution projection operator $\DD$ has been widely studied in the literature and can be assumed to be known.
It consists of band-dependent low-pass filtering followed by spatial decimation at pace $R$
\begin{equation}
    \DD(M_n(\cdot,\cdot,b)) = \left[ M_n(\cdot,\cdot,b) \ast h_b \right] \downarrow R
\label{eq:decimation}
\end{equation}
Under this assumption, $\LL_\lambda$ can be expected to completely vanish in the presence of correct pansharpening, $\wM_0=M_0$,
a property not always satisfied by other quality indicators \cite{Scarpa2021}.

However, this is really the case only if the original spectral bands are correctly aligned, otherwise a co-registration step is required.
Indeed, in multi-resolution imagery, the MS bands are often misaligned.
This is due to technological constraints of the sensing systems and may also depend on the specific product released.
As a result, spectral aberrations appear in the image,
easily spotted in false-color representations as thin lines with weird colors near object boundaries.
Therefore, it is good practice to co-register the MS spectral bands beforehand, a step often neglected by researcher and practitioners alike.
Interestingly, in the proposed framework, bands are automatically co-registering.
In fact, to maximize their spatial correlation with the PAN, they are eventually aligned with it, and hence among themselves.
This good thing, however, has a perverse effect.
After decimation, in fact, the well-aligned low-resolution projection will be compared with a misaligned reference,
generating a non-zero loss even in the presence of a perfect output.
However, this problem is readily solved.
The band-to-PAN shifts resulting after the fine tuning are used in the decimation step to realign $\DD(\wM_0)$ with $M_1$.

In the proposed loss function of eq.(\ref{eq:actual_loss}), two critical hyper-parameters must be set:
the patch size $\sigma$ used in the spatial loss term, and the weight $\beta$ which balances spatial and spectral losses.
In Subsection \ref{sec:parameters} we describe and discuss the preliminary experiments carried out to select the values of $\sigma$ and $\beta$ used in our implementation.

\subsection{Target-adaptive operating modality}

\begin{figure}
\centering
\usetikzlibrary{positioning,shapes,arrows,arrows.meta}
\usetikzlibrary{patterns}
\usetikzlibrary{fit}
\usetikzlibrary{calc,3d}
\definecolor{preCol}{rgb}{0.6, 0.6, 0.6}
\definecolor{adaCol}{rgb}{0.6, 0.6, 0.6}
\definecolor{panCol}{rgb}{0.6, 0.6, 0.6}
\tikzstyle{Circle}=[circle, draw, fill=white, minimum size = 22pt, inner sep=0pt]
\tikzstyle{DoubleCircle}=[circle, double, double distance=1pt, draw, fill=white,minimum size = 20.5pt, inner sep=0pt]
\tikzstyle{Cloud}=[cloud, cloud puffs=15.34, cloud ignores aspect, minimum width=3.5cm, minimum height=2.5cm, align=center, draw]
\tikzstyle{smallCloud}=[cloud, cloud puffs=14.8, cloud ignores aspect, minimum width=2.3cm, minimum height=1.4cm, align=center, draw,inner sep=0pt]
\tikzstyle{Ellipse}=[ultra thick, draw, ellipse, minimum width=6cm, minimum height = 3cm,    align=center]
\tikzstyle{ellip}=[very thick, draw, ellipse, minimum width=2.3cm, minimum height = 1.3cm,    align=center, inner sep = 0pt]
\tikzstyle{ell}=[thick, draw, ellipse, minimum width=2cm, minimum height = 1cm,    align=center, inner sep = 0pt]
\tikzstyle{arr}=[ -stealth, shorten <=3pt, shorten >=3pt]
\tikzstyle{dataArr}=[ -stealth, shorten <=3pt, shorten >=3pt]
\tikzstyle{Box}=[rectangle,  draw, dashed, minimum width=6.6cm, minimum height=4.8cm, inner sep=0pt]
\newcommand{\step}{3}
\scalebox{0.7}{
\begin{tikzpicture}[node distance=2cm]
\node[Cloud](C) at (-0.3*\step,0*\step) {\parbox{1.3cm}{\centering Training dataset}};
\node[smallCloud](c) at (1*\step,0*\step) {};
\node[] at (c) {\parbox{2cm}{\centering Samples \\ from target}};
\node[coordinate](target) at (2.2*\step,0){};
\node[DoubleCircle,xshift = -3mm](P0) at (target) {$P_0$};
\node[DoubleCircle,xshift = 3mm](M1) at (target) {$M_1$};
\node[coordinate, yshift = -3.5mm](target) at (target){};
\node[ellip](train) at (-0.3*\step,-1*\step) {};
\node[ellip](tune) at (1*\step,-1*\step) {};
\node[ellip](pan) at (2.2*\step,-1*\step) {};
\node[] at (train) {Training};
\node[yshift = 2mm] at (tune) {Target};
\node[yshift = -2mm] at (tune) {adaptation};
\node[] at (pan) {Pansharp.};
\node[Circle](out) at (3*\step,-1*\step) {$\widehat{M}_0$};
\node[Box](box) at (1.6*\step,-0.5*\step){};
\node[below of =box, node distance = 2.7cm] {\bf Target-adaptive operation};
\node[coordinate] (n) at (-0.3*\step,-0.5*\step){};
\node[below of =n, node distance = 2.7cm] {\bf Pretraining};
\draw[dataArr] (C)--(train);
\draw[arr] (c)--(tune);
\draw[arr] (P0)--(c);
\draw[arr] (target)--(pan);
\draw[arr] (train)--node[above]{$\phi^{(0)}$}(tune);
\draw[arr] (tune)--node[above]{$\phi^{(\infty)}$}(pan);
\draw[arr] (pan)--(out);
\end{tikzpicture}
}
\caption{High-level flowchart of target-adaptive pansharpening.}
\label{fig:TargetAdaptive}
\end{figure}
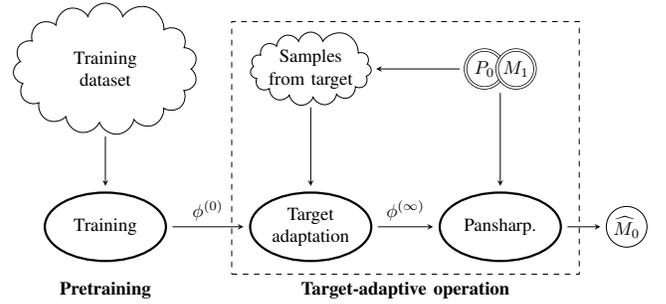

Remote sensing images present a large variability, due to the portrayed scene, the sensor characteristics, the acquisition conditions, etc.
Even a large and well-designed dataset could hardly capture this wide variety,
but the training sets used in practical applications consists often of just one or a few (large) images, often acquired by the same sensor.
This is mostly due to the high cost of multi-resolution images and the scarcity of data freely available for the research community.
Understandably, models trained in these conditions work poorly on new off-training images.
To address this problem, target-adaptive pansharpening was proposed in \cite{Scarpa2018a}.
This operating modality (see Fig.~\ref{fig:TargetAdaptive})
consists in unfreezing the network weights, $\phi^{(0)}$, and performing a few cycles of fine tuning to the target image,
using some selected samples extracted by the target image itself.
With a sensible choice of parameters, only a limited increase in complexity is incurred.
On the positive side, the generalization ability improves sharply,
with performance gains that may be also very significant, depending on training-test mismatch.
We therefore regard target adaptation as an essential ingredient for real-world pansharpening methods
and an integral part of the proposed framework.
At test time, the user is only asked to provide/select the pretrained network,
then the algorithm runs a few iterations of fine tuning to optimize the weights for the target image,
before carrying out the actual pansharpening using the updated parameters, $\phi^{(\infty)}$.
The default number of tuning iterations was set to 50 in \cite{Scarpa2018a}.
Here, we raise it to 100, based on the experimental results discussed in Section~\ref{sec:zpnn}.

\section{Experimental analysis}

\subsection{Reference methods, datasets, performance measures}

\subsubsection{Comparative methods}
For all comparative analyses, we rely on the benchmark toolbox \cite{Vivone2020}
which implements a large number of methods belonging to the four main categories recalled in the introduction: CS, MRA, VO and ML.
All methods available in the toolbox are used in the experiments, except for a few VO solutions which suffer software compatibility issues.
In addition, we consider two more state-of-the-art ML methods, PanNet \cite{Yang2017} and DRPNN \cite{Wei2017L},
retrained on our datasets to ensure a fair comparison.

\subsubsection{Datasets}
Tab.~\ref{tab:datasets} lists the datasets used for training, validation, and fine tuning of the deep learning-based models, and for testing of all methods.
In some cases, we use baseline models pre-trained on other datasets detailed in the reference papers.

\begin{table}
\centering
\begin{tabular}{lccccc}
\hline
\ru    Sensor-site  & \# tiles &         PAN size &  GSD &    Usage   \\ \hline
\ru WV3-Mexico City &        1 & 2048$\times$2048 & 0.31 & Training   \\
\ru WV3-Mexico City &        2 & 2048$\times$2048 & 0.31 & Validation \\
\ru WV3-Adelaide    &       10 & 2048$\times$2048 & 0.31 & Testing    \\
\ru WV2-Napoli      &        1 & 2048$\times$2048 & 0.46 & Training   \\
\ru WV2-Napoli      &        2 & 2048$\times$2048 & 0.46 & Validation \\
\ru WV2-Washington  &       13 & 2048$\times$2048 & 0.46 & Testing    \\
\ru GE1-Waterford   &        1 & 2048$\times$2048 & 0.41 & Training   \\
\ru GE1-Waterford   &        2 & 2048$\times$2048 & 0.41 & Validation \\
\ru GE1-Genova      &       10 & 2048$\times$2048 & 0.41 & Testing    \\ \hline
\end{tabular}
\caption{
Datasets. GSD: PAN ground sample distance at nadir [m].
PAN/MS resolution ratio, $R$=4.
Adelaide and Washington, courtesy of DigitalGlobe$^\copyright$.
Mexico City, Napoli, Waterford and Genova (DigitalGlobe$^\copyright$) provided by ESA.
}
\label{tab:datasets}
\end{table}

\subsubsection{Performance measures}
Assessing the performance of pansharpening methods is an open issue, given the lack of full-resolution ground truths.
A widespread approach is to measure performance objectively in a reduced resolution setting.
Popular indexes used to this end are SAM (Spectral Angle Mapper),
ERGAS ({\em Erreur Relative Globale Adimensionnelle de Synth{\'e}se}), and
$Q2^{n}$ (multiband extension of the Universal Image Quality Index, UIQI) \cite{Wald2002, Alparone2004, Garzelli2009},
also provided in the benchmark toolbox \cite{Vivone2020}.
However, this approach is at odds with our goals, and is not followed here.

Instead, we consider full-resolution no-reference indexes, which assess separately spectral and spatial fidelity.
Many such indexes have been proposed in recent years towards this end, for example \cite{Pradhan2006, Zhou1998, Meng2021}.
For spectral fidelity, we consider here the spectral distortion index, $\DL$, proposed by Khan \cite{Khan2009},
in the slightly modified implementation of the assessment toolbox \cite{Vivone2020},
together with the reprojection indexes, R-SAM, R-ERGAS, and R-$Q2^n$, proposed in \cite{Scarpa2021}.
Note that R-$Q2^n$ equals 1-$\DL$ if the latter is implemented as originally proposed.
For spatial fidelity, instead,
we consider the spatial distortion index, $\DS$, proposed in \cite{Alparone2008}, and the correlation distortion index, $\DR$, also proposed in \cite{Scarpa2021}.
Unlike for the spectral case, these two indexes have a deeply different rationale, and sometimes provide contrasting results.
In particular, experiments carried out in \cite{Scarpa2021} show that
$\DR$ correlates better than $\DS$ with experts' visual assessment, especially for high-quality pansharpening.

\begin{table}
\centering
\begin{tabular}{lc@{\rule{2pt}{0pt}}ccc@{\rule{2pt}{0pt}}c@{\rule{2pt}{0pt}}c}
\hline
\ru      full acronym & \multicolumn{2}{c}{pretraining} & & \multicolumn{3}{c}{target-adaptation} \\ \cline{2-3} \cline{5-7}
\ru                   &            dataset & resolution & &       applied & resolution & \# iter. \\ \hline
\ru {\em model}       &           authors' &    reduced & &            no &         -- &       -- \\ \hline
\ru {\em model}$^*$   &               ours &    reduced & &            no &         -- &       -- \\ \hline
\ru {\em model}-TA    &           authors' &    reduced & &           yes &    reduced &     2000 \\ \hline
\ru {\em model}-TA-FR &           authors' &    reduced & &           yes &       full &     2000 \\ \hline
\ru Z-PNN (0 iter.)   &               ours &       full & &            no &         -- &       -- \\ \hline
\ru Z-PNN             &               ours &       full & &           yes &       full &      100 \\ \hline
\end{tabular}
\caption{For each {\em model} $\in$ \{A-PNN, PanNet, DRPNN\} we consider several versions, differing in pre-training and target adaptation.
Z-PNN is a proposed PNN variant later detailed (Sec.~\ref{sec:zpnn}).}
\label{tab:variants}
\end{table}

\subsection{Does full-resolution training improve performance?}

Aim of this Subsection is to prove that the proposed full-resolution training framework does indeed improve the performance of deep learning-based pansharpening,
as measured by full-resolution quality indexes and especially visual inspection.
Towards this end, we consider three state-of-the-art networks, PanNet \cite{Yang2017}, DRPNN \cite{Wei2017L}, and A-PNN \cite{Scarpa2018a},
a variant of PNN \cite{Masi2016} with a skip connection for residual learning.
For each network, we consider three versions.
First of all, the basic {\em model}, as originally trained by the authors using losses based on $L_1$-norm (A-PNN) or $L_2$-norm (the others).
By doing so, we have a solid starting point, the network optimized by the authors on their own data and available online.
Then, we add two target-adaptive versions, with adaptation carried out at reduced resolution, with the Wald-like approach ({\em model}-TA),
or at full resolution, with the proposed framework ({\em model}-TA-FR).
Unlike in normal operations, where only a few iterations are used to save time,
we use a large number of iterations here, 2000, to ensure a very good adaptation to the target image.
This allows the network to “forget” the initial parameters, removing possible biases due to the different pretraining conditions.
At this point,
differences in performance will depend only on the architecture and, for each architecture, on the use of the low- or high-resolution training framework.
Tab.~\ref{tab:variants} summarizes these models and variants.\footnote{In the table
we also include models used in subsequent experiments, that is, the versions retrained on our datasets (marked by an asterisk),
and Z-PNN, a PNN variant proposed here (Sec.\ref{sec:zpnn}).
We warn the reader that the toolbox \cite{Vivone2020} uses a slightly different acronym
(A-PNN-FT, Advanced PNN with Fine-Tuning) to indicate the reduced-resolution target-adaptive A-PNN \cite{Scarpa2018a} that we name here A-PNN-TA.}

\begin{figure}
\centering
\pgfplotsset{compat=newest}
\usetikzlibrary{plotmarks}
\usetikzlibrary{matrix}
\newcommand{\DLy}{$D_\lambda^{({\rm K})}$}
\definecolor{PREcol}{rgb}{0.85,0.85,0.85}
\definecolor{FTcol}{rgb}{0.6,0.6,0.6}
\definecolor{ZOOMcol}{rgb}{0.2,0.4,0.7}
\definecolor{ZOOMCcol}{rgb}{0.2,0.7,0.4}
\pgfplotstableread{
method pre ft zoom zoomc	
A-PNN            0.1038  0.0779  0.0460  0.1003              
PanNet           0.1189  0.0602  0.0356 0.1036              
DRPNN            0.1649  0.1144  0.0581 0.1246              
}\mytableDL 
\pgfplotstableread{
method pre ft zoom zoomc	
A-PNN        0.0950  0.0441  0.0716  0.0878           
PanNet       0.0708  0.0392  0.0823 0.0762            
DRPNN        0.0643  0.0347  0.0933 0.0866         
}\mytableDS
\pgfplotstableread{
method pre ft zoom 	zoomc
A-PNN       0.8122  0.8817  0.8859  0.8208   
PanNet      0.8189  0.9030  0.8850  0.8280  
DRPNN       0.7812  0.8550  0.8542  0.7996   
}\mytableHQNR 
\pgfplotstableread{
method pre ft zoom 	zoomc
A-PNN    0.5355  0.3331  0.0911  0.0840   
PanNet   0.5997  0.4700  0.0705  0.0581 
DRPNN    0.2562  0.2642  0.0997  0.0891 
}\mytableRHO 
\pgfplotstableread{
method pre ft zoom 	zoomc
A-PNN 				0.9540  0.9648  0.9753  0.9412
PanNet				0.9407  0.9692  0.9808  0.9395
DRPNN				0.9212  0.9393  0.9720  0.9310 
}\mytableQ
\pgfplotstableread{
method pre ft zoom 	zoomc
A-PNN 				2.6297  2.7792  2.7699  4.3297
PanNet				3.8428  2.9217  2.3159  4.2335 
DRPNN				5.5856  3.5921  2.8766  4.4955 
}\mytableSAM
\pgfplotstableread{
method pre ft zoom 	zoomc
A-PNN 				2.5648 2.5005  2.1697  3.6103
PanNet				3.1983 2.5052  1.9778  3.5770 
DRPNN				3.9277 3.4004  2.2070  3.7292
}\mytableERGAS
\pgfplotstableread{
method pre ft zoom 	zoomc
A-PNN 				0.9307  0.9422  0.9538  0.9781	
PanNet				0.9351  0.9638  0.9503  0.9697		
DRPNN				0.9212  0.9440  0.9469  0.9682		
}\mytableQc 
\pgfplotstableread{
method pre ft zoom 	zoomc
A-PNN 				4.1769  4.2429  4.1081  2.7591		
PanNet				4.2610  3.5548  4.1945  3.2834		
DRPNN				5.6721  3.6534  4.2212  3.2317	
}\mytableSAMc 
\pgfplotstableread{
method pre ft zoom 	zoomc
A-PNN 				3.4923  3.3385  3.1455  1.9638
PanNet				3.3570  2.7001  3.3154  2.3843
DRPNN				3.8945  3.2179  3.3107  2.3055
}\mytableERGASc 
\pgfplotsset{
    every axis/.append style={
		width=8cm,     height=6cm,    ybar,
	    enlarge x limits=0.2,
	    legend style={at={(0.5,-0.2)}, anchor = north,legend columns=-1},
	    	ylabel style={at={(0,1)}, anchor=north west, rotate=-90},
	    	symbolic x coords={A-PNN,PanNet,DRPNN},
    		xtick=data,
    		nodes near coords align={vertical},
    }
}
\scalebox{0.55}{
\begin{tikzpicture}
\begin{axis}[ymin=0, ymax=0.2,  ytick={0,0.1,0.2}, ylabel={\DLy},
					name=DL]
	\addplot[fill=PREcol] table[x=method, y=pre] {\mytableDL}; 
	\addplot[fill=FTcol] table[x=method, y=ft] {\mytableDL};
	\addplot[fill=ZOOMcol] table[x=method, y=zoom] {\mytableDL};	
	\end{axis}

\begin{axis}[ymin=0.9, ymax=1, ylabel={R-$Q2^n$},
					name=Q, at=(DL.right of south east), anchor=left of south west]
	\addplot[fill=PREcol] table[x=method, y=pre] {\mytableQ};
	\addplot[fill=FTcol] table[x=method, y=ft] {\mytableQ};
	\addplot[fill=ZOOMcol] table[x=method, y=zoom] {\mytableQ};
\end{axis}

\begin{axis}[ymin=0, ymax=6, ylabel={R-SAM},
					name=SAM, at=(DL.south east), anchor=north east, yshift=-10mm]
	\addplot[fill=PREcol] table[x=method, y=pre] {\mytableSAM};
	\addplot[fill=FTcol] table[x=method, y=ft] {\mytableSAM};
	\addplot[fill=ZOOMcol] table[x=method, y=zoom] {\mytableSAM};
\end{axis}

\begin{axis}[ymin=0, ymax=5, ylabel={R-ERGAS},
					name=ERGAS, at=(Q.south east), anchor=north east, yshift=-10mm]
	\addplot[fill=PREcol] table[x=method, y=pre] {\mytableERGAS};
	\addplot[fill=FTcol] table[x=method, y=ft] {\mytableERGAS};
	\addplot[fill=ZOOMcol] table[x=method, y=zoom] {\mytableERGAS};
	\legend{Pretrained,TA, TA-FR};
\end{axis}
\end{tikzpicture}
}
\caption{Full-resolution {\em spectral} accuracy indexes for Adelaide.}
\label{fig:histSpectralFR}
\end{figure}
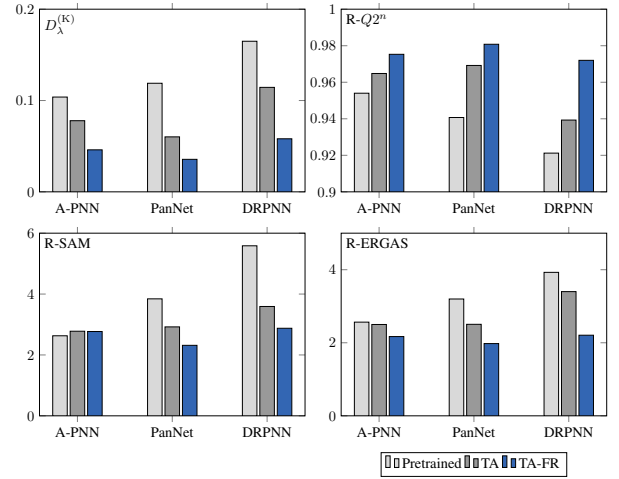

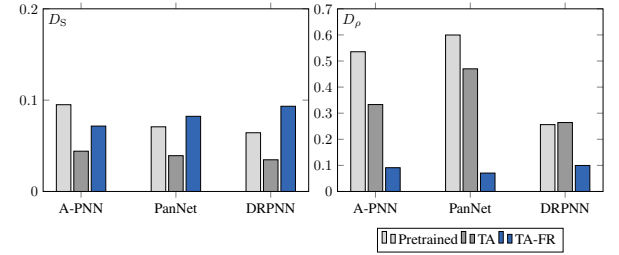
\begin{figure}
\centering
\pgfplotsset{compat=newest}
\usetikzlibrary{plotmarks}
\usetikzlibrary{matrix}
\newcommand{\DLy}{$D_\lambda^{({\rm K})}$}
\definecolor{PREcol}{rgb}{0.85,0.85,0.85}
\definecolor{FTcol}{rgb}{0.6,0.6,0.6}
\definecolor{ZOOMcol}{rgb}{0.2,0.4,0.7}
\definecolor{ZOOMCcol}{rgb}{0.2,0.7,0.4}
\pgfplotstableread{
method pre ft zoom zoomc	
A-PNN            0.1038  0.0779  0.0460  0.1003              
PanNet           0.1189  0.0602  0.0356 0.1036              
DRPNN            0.1649  0.1144  0.0581 0.1246              
}\mytableDL 
\pgfplotstableread{
method pre ft zoom zoomc	
A-PNN        0.0950  0.0441  0.0716  0.0878           
PanNet       0.0708  0.0392  0.0823 0.0762            
DRPNN        0.0643  0.0347  0.0933 0.0866         
}\mytableDS
\pgfplotstableread{
method pre ft zoom 	zoomc
A-PNN       0.8122  0.8817  0.8859  0.8208   
PanNet      0.8189  0.9030  0.8850  0.8280  
DRPNN       0.7812  0.8550  0.8542  0.7996   
}\mytableHQNR 
\pgfplotstableread{
method pre ft zoom 	zoomc
A-PNN    0.5355  0.3331  0.0911  0.0840   
PanNet   0.5997  0.4700  0.0705  0.0581 
DRPNN    0.2562  0.2642  0.0997  0.0891 
}\mytableRHO 
\pgfplotstableread{
method pre ft zoom 	zoomc
A-PNN 				0.9540  0.9648  0.9753  0.9412
PanNet				0.9407  0.9692  0.9808  0.9395
DRPNN				0.9212  0.9393  0.9720  0.9310 
}\mytableQ
\pgfplotstableread{
method pre ft zoom 	zoomc
A-PNN 				2.6297  2.7792  2.7699  4.3297
PanNet				3.8428  2.9217  2.3159  4.2335 
DRPNN				5.5856  3.5921  2.8766  4.4955 
}\mytableSAM
\pgfplotstableread{
method pre ft zoom 	zoomc
A-PNN 				2.5648 2.5005  2.1697  3.6103
PanNet				3.1983 2.5052  1.9778  3.5770 
DRPNN				3.9277 3.4004  2.2070  3.7292
}\mytableERGAS
\pgfplotstableread{
method pre ft zoom 	zoomc
A-PNN 				0.9307  0.9422  0.9538  0.9781	
PanNet				0.9351  0.9638  0.9503  0.9697		
DRPNN				0.9212  0.9440  0.9469  0.9682		
}\mytableQc 
\pgfplotstableread{
method pre ft zoom 	zoomc
A-PNN 				4.1769  4.2429  4.1081  2.7591		
PanNet				4.2610  3.5548  4.1945  3.2834		
DRPNN				5.6721  3.6534  4.2212  3.2317	
}\mytableSAMc 
\pgfplotstableread{
method pre ft zoom 	zoomc
A-PNN 				3.4923  3.3385  3.1455  1.9638
PanNet				3.3570  2.7001  3.3154  2.3843
DRPNN				3.8945  3.2179  3.3107  2.3055
}\mytableERGASc 
\pgfplotsset{
    every axis/.append style={
		width=8cm,     height=6cm,    ybar,
	    enlarge x limits=0.2,
	    legend style={at={(0.5,-0.2)}, anchor = north,legend columns=-1},
	    	ylabel style={at={(0,1)}, anchor=north west, rotate=-90},
	    	symbolic x coords={A-PNN,PanNet,DRPNN},
    		xtick=data,
    		nodes near coords align={vertical},
    }
}
\newcommand{\DSy}{$D_{\rm S}$}
\newcommand{\DRy}{$D_\rho$}
\scalebox{0.55}{
\begin{tikzpicture}
\begin{axis}[ymin=0, ymax=0.2,  ytick={0,0.1,0.2}, ylabel={\DSy},
					name=DS]
	\addplot[fill=PREcol] table[x=method, y=pre] {\mytableDS}; 
	\addplot[fill=FTcol] table[x=method, y=ft] {\mytableDS};
	\addplot[fill=ZOOMcol] table[x=method, y=zoom] {\mytableDS};	
	\end{axis}

\begin{axis}[ymin=0, ymax=0.7, ytick={0,0.1,0.2,0.3,0.4,0.5,0.6,0.7}, ylabel={\DRy},
					name=DR, at=(DS.right of south east), anchor=left of south west]
	\addplot[fill=PREcol] table[x=method, y=pre] {\mytableRHO};
	\addplot[fill=FTcol] table[x=method, y=ft] {\mytableRHO};
	\addplot[fill=ZOOMcol] table[x=method, y=zoom] {\mytableRHO};
	\legend{Pretrained,TA, TA-FR};
\end{axis}
\end{tikzpicture}
}
\caption{Full-resolution {\em spatial} accuracy indexes for Adelaide.}
\label{fig:histSpatialFR}
\end{figure}

Fig.~\ref{fig:histSpectralFR} and Fig.~\ref{fig:histSpatialFR} report spectral and spatial quality indexes, respectively,
obtained for the WorldView-3 Adelaide test image.
Similar results are obtained with different test images.

A first observation concerns the significant performance gaps observed between different pretrained models (light gray bins).
For example, A-PNN has an R-SAM index about half that of DRPNN.
However, such differences depend more on the limited generalization ability of the methods than on their intrinsic effectiveness.
A-PNN was originally trained on data well-aligned with our WorldView-3 test image, something that probably did not happen with DRPNN.
This interpretation is strongly supported by results obtained with the TA models (dark gray bins).
In fact, with target adaptation the performance improves almost always significantly,
and the quality indexes become much more uniform across the three methods.
Overall, target adaptation mitigates the mismatch between training and test set
and the resulting indexes can be regarded as more reliable indicators of the actual potential of the various pansharpening tools.

\newcommand{\mybox}[1]{{\framebox{\parbox{1.4cm}{\centering \rule{0cm}{2.1cm}#1}}}}
\newcommand{\pathA}{./figures/Fig07JPG/}
\newcommand{\imA}[1]{\includegraphics[width=1.6cm]{\pathA #1.jpg}}
\begin{figure}
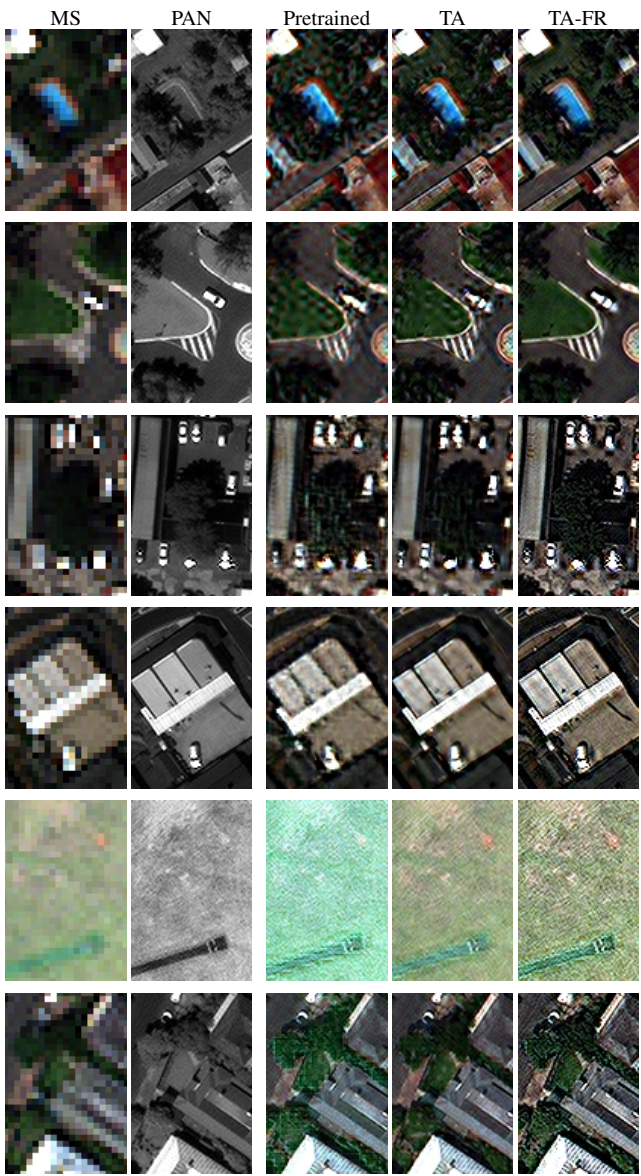

\centering
\tiny
\setlength{\tabcolsep}{1pt}
\begin{tabular}{cc@{\rule{2mm}{0mm}}ccc}
\footnotesize MS & \footnotesize PAN & \footnotesize Pretrained &  \footnotesize TA & \footnotesize TA-FR \\ \vspace{1mm}
\imA{MS_1} & \imA{PAN_1} & \imA{APNN_1} & \imA{APNN_FT_1} & \imA{ZPNN_1} \\ \vspace{1mm}
\imA{MS_2} & \imA{PAN_2} & \imA{APNN_2} & \imA{APNN_FT_2} & \imA{ZPNN_2} \\ \vspace{1mm}
\imA{MS_3} & \imA{PAN_3} & \imA{Pannet_3} & \imA{Pannet_FT_3} & \imA{ZPannet_3} \\ \vspace{1mm}
\imA{MS_4} & \imA{PAN_4} & \imA{Pannet_4} & \imA{Pannet_FT_4} & \imA{ZPannet_4} \\ \vspace{1mm}
\imA{MS_5} & \imA{PAN_5} & \imA{DRPNN_5} & \imA{DRPNN_FT_5} & \imA{ZDRPNN_5} \\ \vspace{1mm}
\imA{MS_6} & \imA{PAN_6} & \imA{DRPNN_6} & \imA{DRPNN_FT_6} & \imA{ZDRPNN_6} \\
\end{tabular}
\caption{Results on crops from the Adelaide image.
From left to right: MS, PAN, Pretrained, TA, TA-FR.
From top to bottom: A-PNN (rows 1-2), PanNet (rows 3-4), DRPNN (rows 5-6).
Red, Green and Blue bands are used for RGB composition.}
\label{fig:cropsFrameworkFR}
\end{figure}

We now turn to the real objective of our analysis,
the performance obtained with target adaptation at high resolution (blue) to be compared with that obtained at low resolution (dark gray).
Regarding spectral quality,
a significant gain is observed for all methods over all indexes (again, with minor exceptions), and the performance appears to be even more uniform than before.
For spatial quality, instead, results are more controversial.
While the $\DR$ index drops, suggesting a large quality improvement, the $\DS$ index grows again, indicating a spatial accuracy comparable to that of pretrained models.
Two facts motivate this strong mismatch.
On one hand, we argue that $\DS$ is not really a reliable indicator when quality is very high.
Indeed, as also noted in \cite{Scarpa2021}, $\DS$ does not really measure spatial quality, but rather a sort of cross-scale spatial quality consistency.
So, it may be small even in the presence of strong spatial distortion, provided the same distortion occurs across the various scales of interest,
and it may be large even with perfect pansharpening, $\wM_0=M_0$.
On the other hand, since the spatial loss used in our training framework follows closely the definition of $\DR$, this indicator may be biased in favor of TA-FR methods.
Since such contradictions cannot be reconciled, we will keep using both indicators, leaving the final say to visual inspection.

In Fig.~\ref{fig:cropsFrameworkFR},
for some crops of the Adelaide test image,
we show the original MS and PAN data together with the output pansharpened images obtained with the pretrained, TA, and TA-FR versions of the three CNN-based methods.
Since we are interested in comparing training schemes against one another, not architectures,
we show different crops for different architectures so as to offer a richer yet compact picture.
First of all, visual inspection fully confirms the improvements in terms of spectral accuracy ensured by target adaptation.
With respect to pretrained models, colors are better preserved and some evident errors are avoided.
In addition, the TA-FR solutions seem to ensure clear improvements also in terms of spatial accuracy.
Some strange patterns created by pretrained or TA networks disappear.
Small objects ({\it e.g.}, the cars) are reconstructed with higher accuracy and, in general, all contours are sharper.
High-frequency textures observed in the PAN are preserved (sometimes, even oversharpened).
Overall, we see a huge improvement with respect to the pretrained models, as predicted by $\DR$,
and also a consistent improvement with respect to the TA versions.
While further work is certainly necessary to obtain fully satisfactory pansharpening,
we believe that these results represent convincing indications that high-resolution training is the right path to follow.

\begin{table}
\setlength{\tabcolsep}{1mm}
\centering
\begin{tabular}{l@{\rule{5mm}{0mm}}rrr@{\rule{5mm}{0mm}}rrr}
\hline
\ru & \multicolumn{3}{c}{Time [seconds]} & \multicolumn{3}{c}{GPU Memory [GB]} \\
\ru & A-PNN & PanNet & DRPNN & A-PNN & PanNet & DRPNN \\
\hline
\ru TA		& 0.862 & 0.346 &  0.615 & 0.38 &  0.74 &  2.03 \\
\ru TA-FR    & 1.885 & 1.885 & 11.172 & 8.96 & 12.84 & 24.09 \\
\hline
\end{tabular}
\caption{Computation time (per iteration) and memory requirements to perform target adaptation on a 2048$\times$2048 WV3 image.}
\label{tab:costs}
\end{table}

\begin{figure}
\centering
\includegraphics[width=0.98\columnwidth]{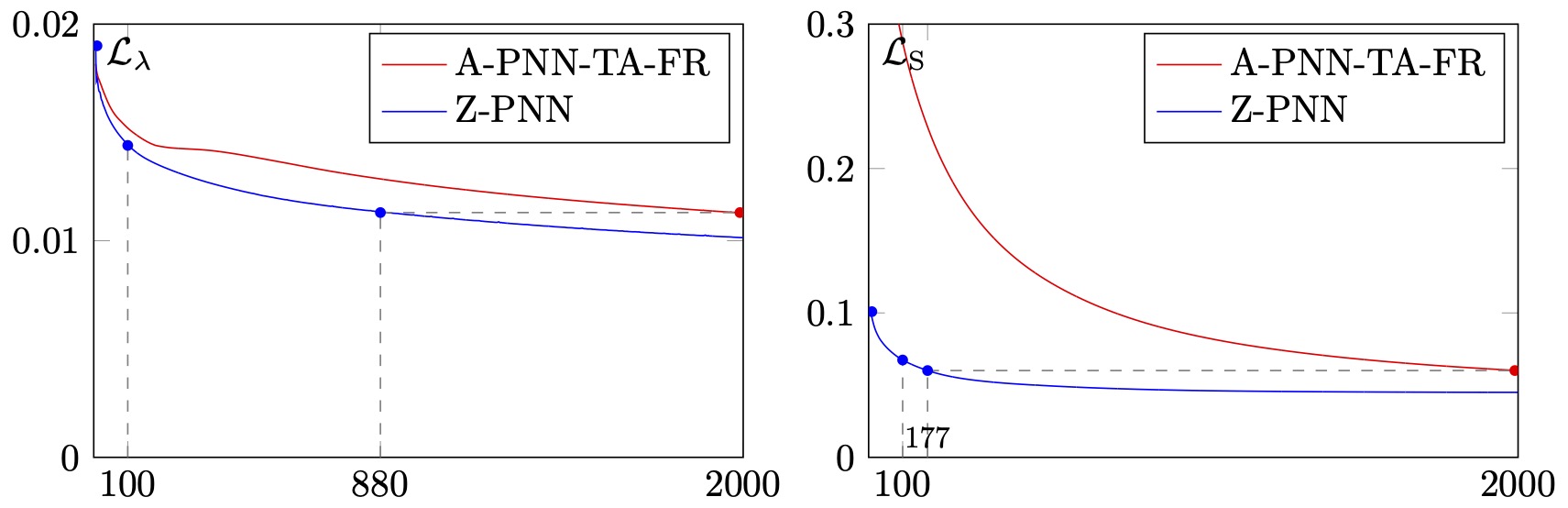}
\caption{Spectral (left) and spatial (right) losses vs. number of iterations for adapting A-PNN-TA-FR (red lines) and Z-PNN (blue lines) to the target image.}
\label{fig:loss}
\end{figure}

\subsection{Z-PNN: a CNN-based pansharpening method pretrained at full resolution}
\label{sec:zpnn}

The analysis of previous Subsection sheds light on the potential of high-resolution training.
However, it relies on intensive target adaptation, which has non-negligible costs in terms of both memory and time.
Such costs are summarized in Tab.~\ref{tab:costs} for the three models analyzed thus far, considering a 2048$\times$2048-pixel multi-resolution image and a NVIDIA Quadro P6000 GPU.
In practice, 2000 iterations require one hour processing time or more.
This was not the case with the target-adaptive method proposed in \cite{Scarpa2018a}, as it worked on much smaller ($16\times$) low-resolution images and used only 50 iterations.

\newcommand{\pathFocus}{./figures/Fig09JPG/}
\newcommand{\imFocus}[1]{\includegraphics[width=1.65cm]{\pathFocus #1.jpg}}
\begin{figure}
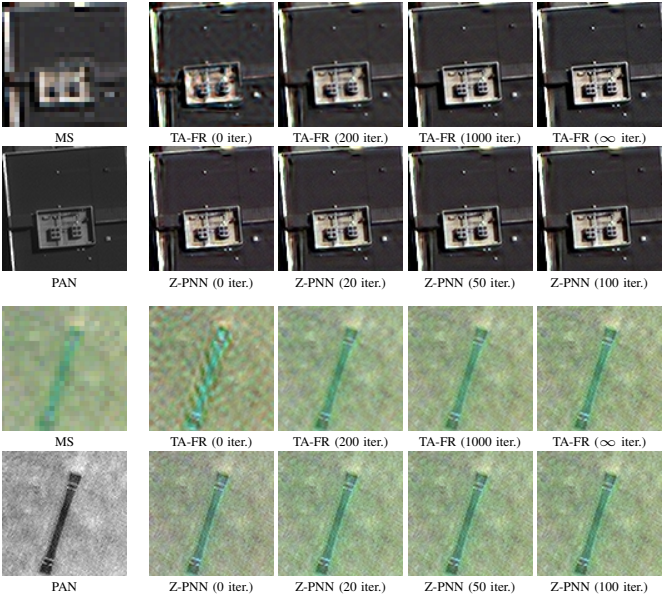

\setlength{\tabcolsep}{1pt}
\centering
\tiny
\begin{tabular}{c@{\rule{3mm}{0mm}}cccc}
\imFocus{MS_1}  & \imFocus{old_0_1} & \imFocus{old_200_1} & \imFocus{old_1000_1} &    \imFocus{old_2000_1} \\
             MS & TA-FR (0 iter.) & TA-FR (200 iter.) & TA-FR (1000 iter.) & TA-FR ($\infty$ iter.)\\
\imFocus{PAN_1} & \imFocus{new_0_1} &  \imFocus{new_20_1} &   \imFocus{new_50_1} &     \imFocus{new_100_1} \\
            PAN &   Z-PNN (0 iter.) &    Z-PNN (20 iter.) &     Z-PNN (50 iter.) &       Z-PNN (100 iter.) \\[2mm]
\imFocus{MS_2}  & \imFocus{old_0_2} & \imFocus{old_200_2} & \imFocus{old_1000_2} &    \imFocus{old_2000_2} \\
             MS & TA-FR (0 iter.) & TA-FR (200 iter.) & TA-FR (1000 iter.) & TA-FR ($\infty$ iter.)\\
\imFocus{PAN_2} & \imFocus{new_0_2} &  \imFocus{new_20_2} &   \imFocus{new_50_2} &     \imFocus{new_100_2} \\
            PAN &   Z-PNN (0 iter.) &    Z-PNN (20 iter.) &     Z-PNN (50 iter.) &       Z-PNN (100 iter.)
\end{tabular}
\caption{Impact of target adaptation with increasing iterations on image quality for (A-PNN-)TA-FR (odd rows) and Z-PNN (even rows) for two WV3 crops.
MS and PAN are shown on the left.
Z-PNN reaches a satisfactory quality long before A-PNN-TA-FR.
}
\label{fig:zpnn}
\end{figure}

To obtain {\em fast} high-quality pansharpening,
we refined the original model weights through a further pretraining phase carried out at full resolution,
using a dedicated training image for each sensor (see again Tab.\ref{tab:datasets}).
By doing so, we expect that much fewer iterations will be necessary for target adaptation.
Since all three architectures appear to perform equally well, from now on we focus only on the simplest one, A-PNN.
The resulting network will be referred to as Z-PNN, short for Zoom-PNN.
We test the impact of this modification on an off-training WorldView-3 image.
Fig.~\ref{fig:loss} shows the progress of spectral and spatial loss terms, as adaptation proceeds, for the versions without (A-PNN-TA-FR) and with (Z-PNN) this further pretraining phase.
The right part, concerning the spatial loss, is especially telling.
While the A-PNN-TA-FR curve lowers very slowly, reaching eventually the value ${\cal L}_S \simeq 0.06$ after 2000 iterations,
the Z-PNN curve reaches the same value after less than 200 iterations.
Actually, the spatial loss is quite low from the beginning, ${\cal L}_S \simeq 0.10$, ensuring a good performance even in the absence of any adaptation.
The left figure, instead, shows that the spectral loss benefits from fine tuning also when starting from the Z-PNN weights.
In any case, a small number of iterations seems to be sufficient to observe a significant improvement.
Fig.~\ref{fig:zpnn} shows, for two crops of the test image, the evolution of the pansharpened output as adaptation goes on.
The images fully confirm all previous observations.
In summary, it appears that Z-PNN could be safely used even without adaptation, or with just a few iterations.
In the following experiments, we set conservatively the number of iterations to 100.
However, the user is free to change this value depending on both available resources and quality target.

\subsection{Comparative analysis}

We can now move to a full-fledged comparative analysis.
Experiments will be carried out on test images acquired by three different sensors (WV2, WV3, and GE1), listed in Tab.~\ref{tab:datasets},
and results will be compared with those of the reference methods summarized in Tab.\ref{tab:legend}.

\begin{figure}
\centering
\includegraphics[width=0.945\columnwidth]{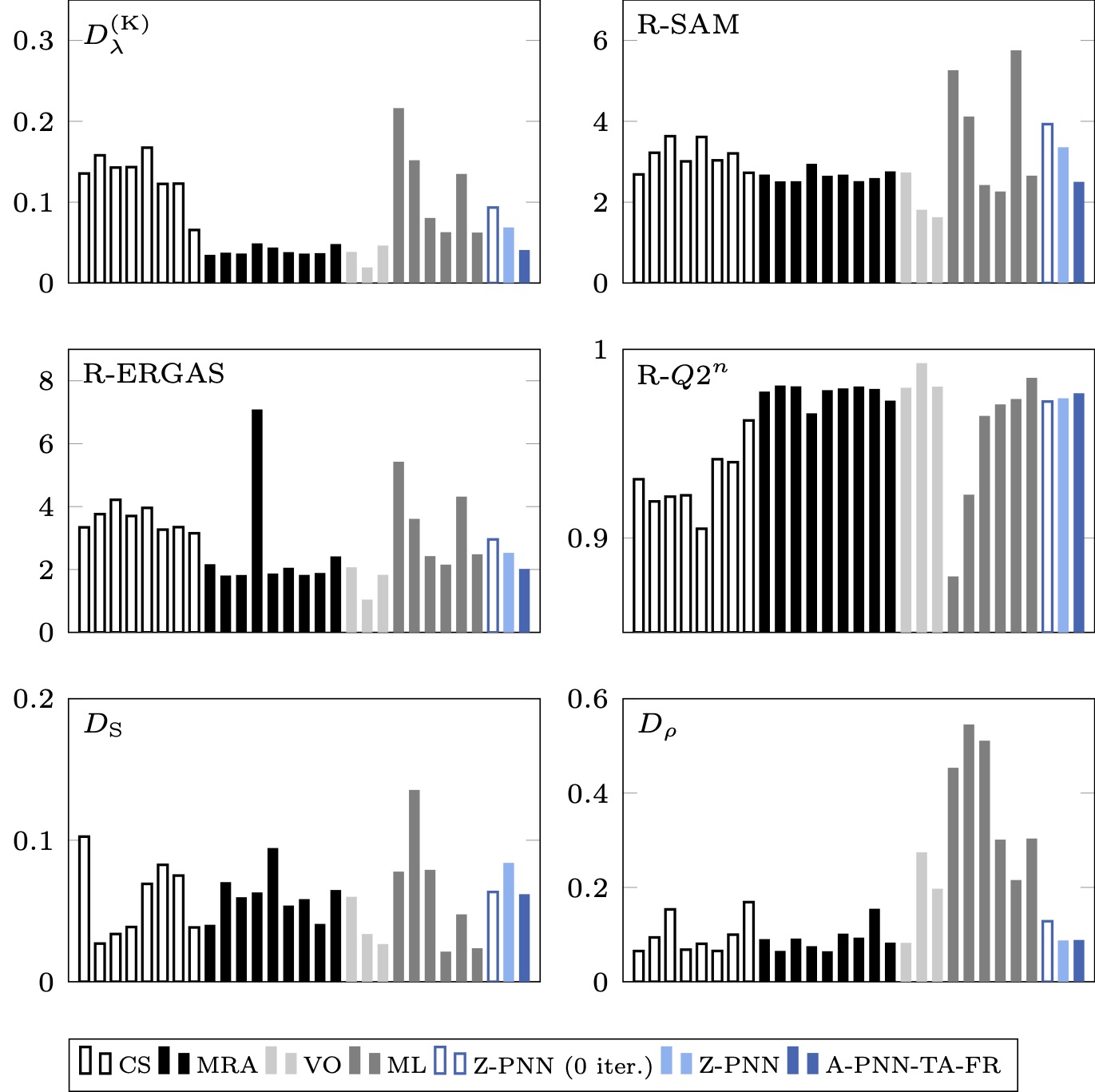}
\caption{Numerical results on the WV3-Adelaide dataset.}
\label{fig:WV3FRcomparison}
\end{figure}

\begin{figure}
\centering
\includegraphics[width=0.945\columnwidth]{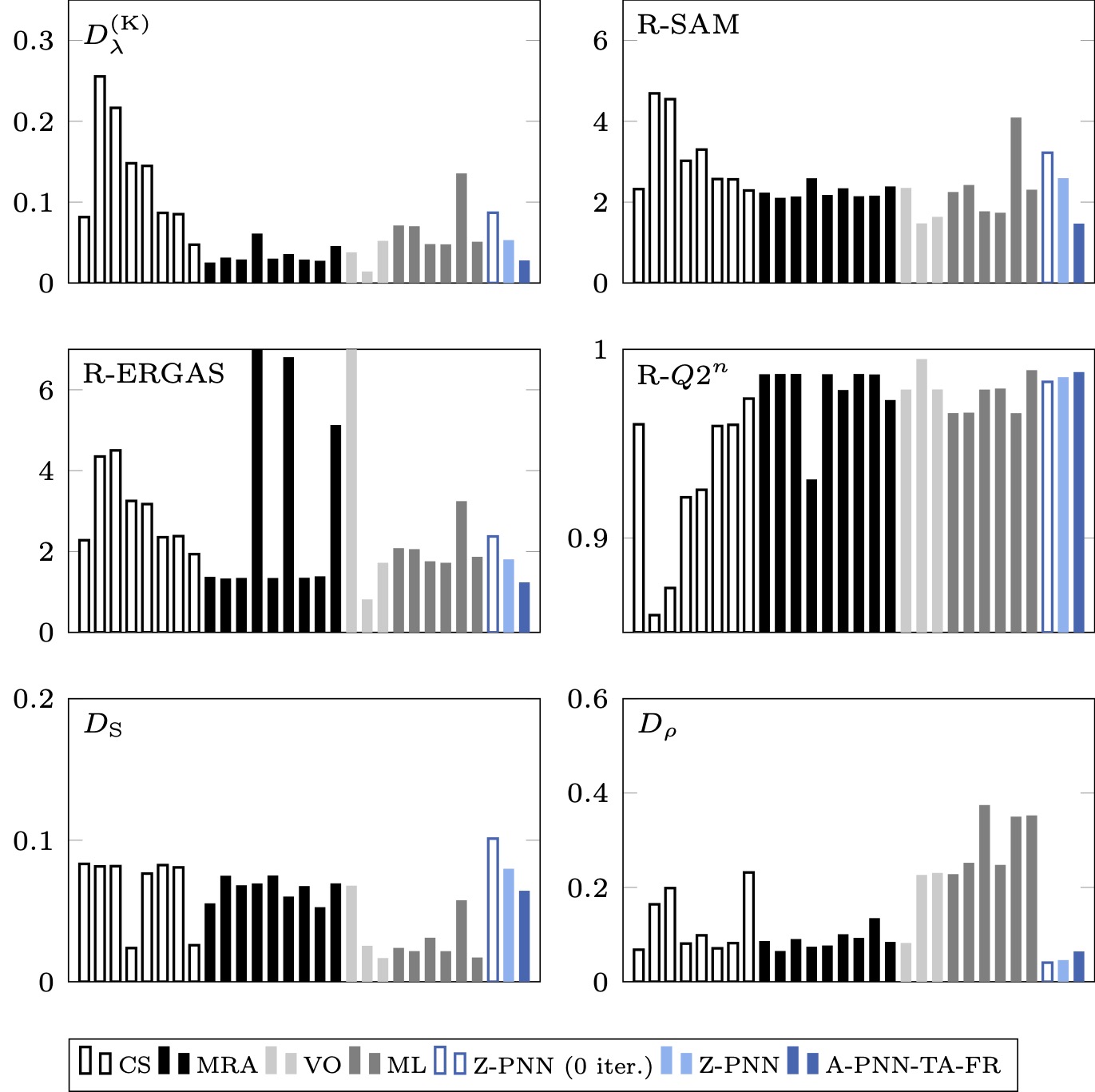}
\caption{Numerical results on the WV2-Washington dataset.}
\label{fig:WV2FRcomparison}
\end{figure}

\begin{figure}
\centering
\includegraphics[width=0.945\columnwidth]{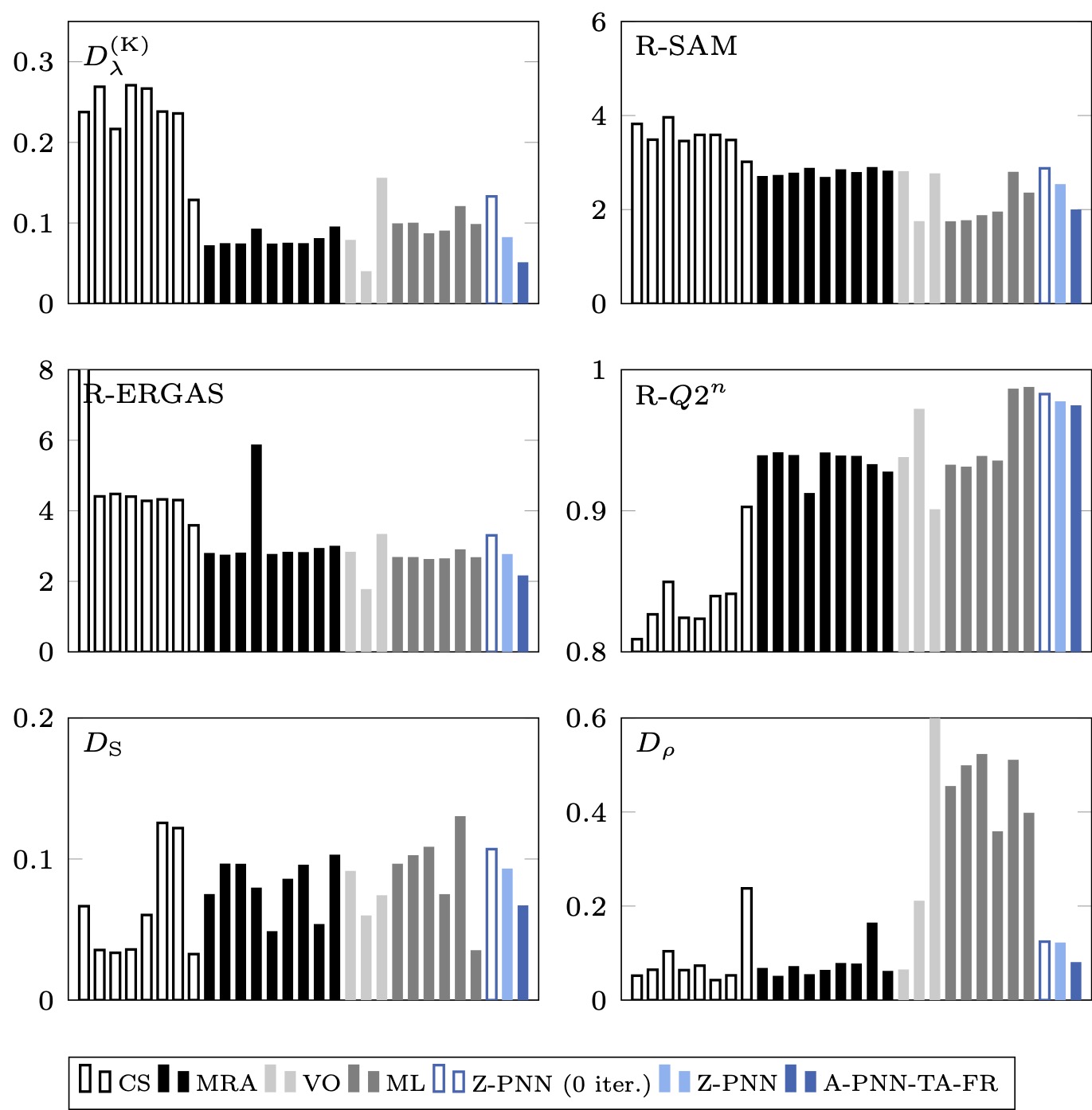}
\caption{Numerical results on the GE1-Amsterdam dataset}
\label{fig:GE1FRcomparison}
\end{figure}

Numerical results are shown in the bar graphs of Figures \ref{fig:WV3FRcomparison}, \ref{fig:WV2FRcomparison} and \ref{fig:GE1FRcomparison},
for the WV3, WV2 and GE1 datasets, respectively.
Each graph refers to a different full-resolution measure, and each bar to a different method.
The reference methods are those listed in Tab.~\ref{tab:legend}, grouped according to their approach (CS, MRA, VO, ML), and shown with a different bar style for each group.
Newly developed methods, Z-PNN with (100 iterations) and without (0) target adaptation, and A-PNN-TA-FR, are shown in shades of blue at the end.
Note that PanNet and DRPNN have been re-trained on our dataset to ensure a fairer comparison, an asterisk marks this version.

\begin{table}
\centering
\setlength{\tabcolsep}{2pt}
\begin{tabular}{p{8.4cm}}
\hline
\multicolumn{1}{c}{\ru \bf Component Substitution (CS)} \\
\ru BT-H \cite{Lolli2017}, BDSD \cite{Garzelli2008}, C-BDSD \cite{Garzelli2015}, BDSD-PC \cite{Vivone2019}, GS \cite{Laben2000},\\
	GSA \cite{Aiazzi2007}, C-GSA \cite{Restaino2017}, PRACS \cite{Choi2011} \\ \hline
\multicolumn{1}{c}{\ru \bf Multiresolution Analysis (MRA)} \\
    \ru AWLP \cite{Alparone2017}, MTF-GLP \cite{Alparone2017}, MTF-GLP-FS \cite{Vivone2018a}, MTF-GLP-HPM \cite{Alparone2017},\\
    MTF-GLP-HPM-H \cite{Lolli2017}, MTF-GLP-HPM-R \cite{Vivone2018}, MTF-GLP-CBD \cite{Alparone2007},\\
    C-MTF-GLP-CBD \cite{Restaino2017}, MF \cite{Restaino2016} \\ \hline
\multicolumn{1}{c}{\ru \bf Variational Optimization (VO)} \\
\ru FE-HPM \cite{Vivone2015a}, SR-D \cite{Vicinanza2015}, TV  \cite{Palsson2014} \\ \hline
\multicolumn{1}{c}{\ru \bf Machine Learning (ML)} \\
\ru PNN \cite{Masi2016}, PNN-IDX \cite{Masi2016}, A-PNN \cite{Scarpa2018a}, A-PNN-TA \cite{Scarpa2018a}, DRPNN* \cite{Wei2017L}, PanNet* \cite{Yang2017} \\ \hline
\end{tabular}
\caption{Detailed list of all reference methods.}
\label{tab:legend}
\end{table}

To begin, let us focus on Fig.\ref{fig:WV3FRcomparison}, also because similar considerations, with minor differences, hold for the other cases.
The most notable outcome is that, contrary to the widespread belief, ML methods do not outperform conventional methods (smaller is better for all measures).
As an example, the MRA methods (black) are generally\footnote{Note that individual methods show occasional failures on some images, we neglect these special cases in this high-level analysis.}
superior to ML methods (dark gray) in terms of both spectral quality and spatial quality indicators.
This surprising result is due, in our opinion, to the low-resolution vs. high-resolution mismatch.
ML methods are usually trained at low resolution, with the Wald-like protocol of Fig.\ref{fig:wald}(left), and then tested again with the Wald protocol.
Therefore, it is not surprising that numerical results speak largely in their favor.
Visual analyses on full resolution data, however, have always casted some shadows on the superiority of ML methods.
Such doubts are confirmed here, where results are computed only in terms of {\em high-resolution} indexes.
These provide a more unbiased assessment of performance
and are better predictors of the quality of pansharpened images the end users can expect.

The performance of ML methods improves significantly only when training takes place at high resolution, as proved by the last three (blue) bars.
This behavior is observed, more or less pronounced, with all sensors, see Fig.\ref{fig:WV2FRcomparison} and Fig.\ref{fig:GE1FRcomparison},
and the improvement is especially significant in terms of spatial quality, according to the $\DR$ indicator.
In particular, the fully (2000 iterations) adapted method, A-PNN-TA-FR (last bar), has one of the smallest $\DR$ values
consistently on all datasets.
Moreover, it has also very good spectral quality indicators, suggesting an excellent overall performance.
On the other hand, it is fair to underline that $\DS$ results depict a very different situation, almost opposite to $\DR$.
Again, this calls for accurate visual inspection of pansharpened images, which is the next step of our analysis.

\newcommand{\pathWV}{./figures/Fig13JPG/}
\newcommand{\imWV}[1]{\includegraphics[width=1.75cm]{\pathWV #1.jpg}}
\newcommand{\sps} [1]{{\scriptsize #1}}
\newcommand{\spsa}[1]{{\scriptsize Best #1}}
\newcommand{\spsb}[1]{{\scriptsize 2nd best #1}}
\begin{figure*}
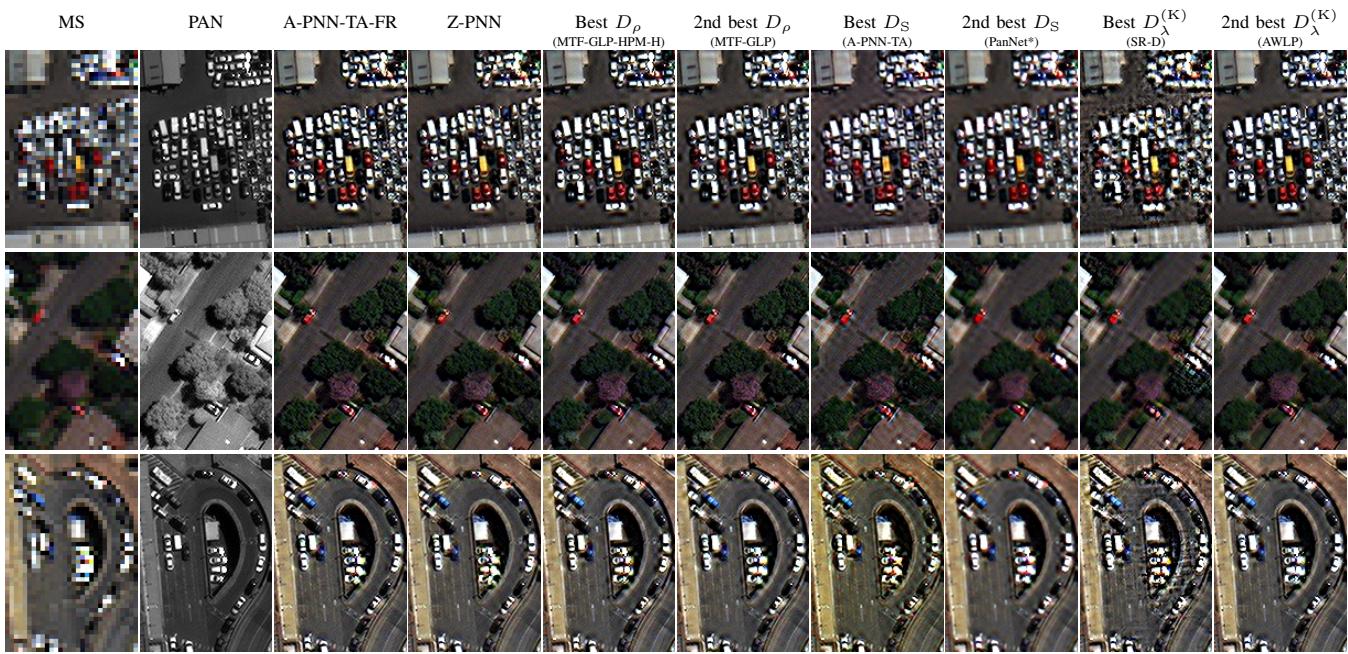

\centering
\tiny
\setlength{\tabcolsep}{0.5pt}
\begin{tabular}{cc@{\rule{1pt}{0pt}}c@{\rule{1pt}{0pt}}ccccccc}
      \sps{MS} &    \sps{PAN} & \sps{A-PNN-TA-FR} &       \sps{Z-PNN} &           \spsa{$\DR$} &     \spsb{$\DR$} &     \spsa{$\DS$} & \spsb{$\DS$} &  \spsa{$\DL$} & \spsb{$\DL$}  \\
               &              &                     &                   &        (MTF-GLP-HPM-H) &        (MTF-GLP) &       (A-PNN-TA) &         (PanNet*) &        (SR-D) &        (AWLP) \\
   \imWV{MS_1} & \imWV{PAN_1} &       \imWV{A-PNN-Zoom-FT_1} & \imWV{Z-PNN_1} & \imWV{MTF-GLP-HPM-H_1} & \imWV{MTF-GLP_1} & \imWV{A-PNN-FT_1} &  \imWV{PanNet_1} & \imWV{SR-D_1} & \imWV{AWLP_1} \\
   \imWV{MS_2} & \imWV{PAN_2} &       \imWV{A-PNN-Zoom-FT_2} & \imWV{Z-PNN_2} & \imWV{MTF-GLP-HPM-H_2} & \imWV{MTF-GLP_2} & \imWV{A-PNN-FT_2} &  \imWV{PanNet_2} & \imWV{SR-D_2} & \imWV{AWLP_2} \\
   \imWV{MS_3} & \imWV{PAN_3} &       \imWV{A-PNN-Zoom-FT_3} & \imWV{Z-PNN_3} & \imWV{MTF-GLP-HPM-H_3} & \imWV{MTF-GLP_3} & \imWV{A-PNN-FT_3} &  \imWV{PanNet_3} & \imWV{SR-D_3} & \imWV{AWLP_3} \\
    \end{tabular}
\caption{Results for some WV3 crops. Left to right: MS, PAN, A-PNN-TA-FR, Z-PNN, best references for $\DR, \DS, \DL$.}
\label{fig:WV3FRcrops}
\end{figure*}

\newcommand{\pathW}{./figures/Fig14JPG/}
\newcommand{\imW}[1]{\includegraphics[width=1.75cm]{\pathW #1.jpg}}
\begin{figure*}
\centering
\tiny
\setlength{\tabcolsep}{0.5pt}
\begin{tabular}{cccccccccc}
\scriptsize MS & \scriptsize PAN &  \scriptsize A-PNN-TA-FR & \scriptsize Z-PNN  & \scriptsize Best $\DR$ & \scriptsize 2nd best $\DR$ & \scriptsize Best $\DS$ & \scriptsize 2nd best $\DS$ & \scriptsize Best $\DL$ & \scriptsize 2nd best $\DL$  \\
 & & & & (MTF-GLP) & (BT-H) & (TV) & (PanNet*) & (SR-D) & (AWLP)   \\
\imW{MS_3} & \imW{PAN_3} & \imW{A-PNN-Zoom-FT_3} & \imW{Z-PNN_3} & \imW{MTF-GLP_3} & \imW{BT-H_3} & \imW{TV_3} & \imW{PanNet_3} & \imW{SR-D_3} & \imW{AWLP_3} \\
\imW{MS_4} & \imW{PAN_4} & \imW{A-PNN-Zoom-FT_4} & \imW{Z-PNN_4} & \imW{MTF-GLP_4} & \imW{BT-H_4} & \imW{TV_4} & \imW{PanNet_4} & \imW{SR-D_4} & \imW{AWLP_4} \\
\imW{MS_5} & \imW{PAN_5} & \imW{A-PNN-Zoom-FT_5} & \imW{Z-PNN_5} & \imW{MTF-GLP_5} & \imW{BT-H_5} & \imW{TV_5} & \imW{PanNet_5} & \imW{SR-D_5} & \imW{AWLP_5} \\
\end{tabular}
\caption{Results for some WV2 crops. Left to right: MS, PAN, A-PNN-TA-FR, Z-PNN, best references for $\DR, \DS, \DL$.}
\label{fig:WV2FRcrops}
\end{figure*}

\newcommand{\myboxF}[1]{{\framebox{\parbox{1.5cm}{\centering \rule{0cm}{2.1cm}#1}}}}
\newcommand{\pathG}{./figures/Fig15JPG/}
\newcommand{\imG}[1]{\includegraphics[width=1.75cm]{\pathG #1.jpg}}
\begin{figure*}
\centering
\tiny
\setlength{\tabcolsep}{0.5pt}
\begin{tabular}{cccccccccc}
\scriptsize MS & \scriptsize PAN &  \scriptsize A-PNN-TA-FR & \scriptsize Z-PNN  & \scriptsize Best $\DR$ & \scriptsize 2nd best $\DR$ & \scriptsize Best $\DS$ & \scriptsize 2nd best $\DS$ & \scriptsize Best $\DL$ & \scriptsize 2nd best $\DL$  \\
 &  &   &  & (GSA) & (MTF-GLP) & (PRACS) & (C-BDSD) & (SR-D) & (AWLP) \\
\imG{MS_1} & \imG{PAN_1} & \imG{A-PNN-Zoom-FT_1} & \imG{Z-PNN_1} & \imG{GSA_1} & \imG{MTF-GLP_1} & \imG{PRACS_1} & \imG{C-BDSD_1} & \imG{SR-D_1} & \imG{AWLP_1} \\
\imG{MS_2} & \imG{PAN_2} & \imG{A-PNN-Zoom-FT_2} & \imG{Z-PNN_2} & \imG{GSA_2} & \imG{MTF-GLP_2} & \imG{PRACS_2} & \imG{C-BDSD_2} & \imG{SR-D_2} & \imG{AWLP_2} \\
\imG{MS_5} & \imG{PAN_5} & \imG{A-PNN-Zoom-FT_5} & \imG{Z-PNN_5} & \imG{GSA_5} & \imG{MTF-GLP_5} & \imG{PRACS_5} & \imG{C-BDSD_5} & \imG{SR-D_5} & \imG{AWLP_5} \\

\end{tabular}
\caption{Results for some GE1 crops. Left to right: MS, PAN, A-PNN-TA-FR, Z-PNN, best references for $\DR, \DS, \DL$.}
\label{fig:GE1FRcrops}
\end{figure*}

Figures \ref{fig:WV3FRcrops}, \ref{fig:WV2FRcrops}, and \ref{fig:GE1FRcrops},
show visual results for some crops acquired by the WorldView-3, WorldView-2 and GeoEye-1 sensors, respectively.
For each crop, next to the original MS and PAN,
we show the output of two methods trained at high resolution (A-PNN-TA-FR and Z-PNN), together with six reference methods.
The latter are chosen as the best and second best ranking methods in terms of $\DR$, $\DS$, and $\DL$, respectively.

Let us consider Fig.\ref{fig:WV3FRcrops}, first, and let us compare the A-PNN-TA-FR with the original PAN-MS pair.
By suitably enlarging the figure, one can fully appreciate the impressive spatial quality of the result.
All details are faithfully preserved with their original shapes and textures, and no alien pattern is introduced by the pansharpening process.
Spectral quality is also very good, but this property is shared with several other methods.
Z-PNN also provides very good results, we only observe a minor loss of spectral accuracy.
Continuing along the row, MTF-GLP-HPM-H and MTF-GLP are the best reference methods in terms of $\DR$, and in fact we observe a very good spatial fidelity also for them.
Things are very different, instead, for A-PNN-TA and PanNet*, the best methods according to $\DS$.
Besides a reduced precision on object shapes and some loss of resolution, especially for PanNet*,
we observe annoying periodic patterns over the whole scene,
confirming that $\DS$ cannot be considered a fully reliable predictor of spatial fidelity.
Finally, the best methods in terms of $\DL$, SR-D and AWLP, ensure indeed a good spectral quality, although comparable to that of other methods,
but exhibit some problems in terms of spatial fidelity.

Fig.\ref{fig:WV2FRcrops} and Fig.\ref{fig:GE1FRcrops} show similar results for the WV2 and GE1 images.
Beyond minor differences, the same phenomena described before are observed in all cases.
A-PNN-TA-FR and Z-PNN keep providing very good results, especially in terms of spatial quality,
only rarely matched by other methods, typically those performing best in terms of $\DR$.

\subsection{Setting loss hyper-parameters, testing alternative losses}
\label{sec:parameters}

In all previous experiments, we used the loss of eq.(\ref{eq:actual_loss}) with hyper-parameters $\sigma$ and $\beta$ optimized experimentally.
Here, we discuss their impact on the performance and motivate experimentally the values selected in the implementation.
In addition, we test an alternative loss fuction proposed in the literature for use in our framework.

\subsubsection{Setting $\sigma$}
the patch size $\sigma$ is the only critical parameter of the proposed spatial loss.
We already motivated the need to estimate the MS-PAN correlation on a {\em local} as opposed to global scale (small $\sigma$),
thereby limiting long-range spatial dependencies and preserving spectral fidelity.
On the other hand, with a very small value for $\sigma$, the correlation ends up being estimated on just a few points.
Lacking any more precise theoretical guidance,
we carried out a number of experiments on test images with $\sigma$ doubling progressively from 2 to 32.
A sample result is shown in Fig.\ref{fig:sigma}.
With $\sigma=2$, obvious artifact are visible on the roads, in the form of diagonal patterns.
These disappear already with $\sigma=4$ and then $\sigma=8$,
however, for larger values other spectral checkerboard aberrations appear on the building rooftops, like echoes of the existing black separation lines.
We observed a similar behaviour on many more test images, which suggests choosing small values for $\sigma$, between 4 and 8.
Eventually, we set $\sigma$ equal to the resolution ratio, $R$, which is always 4 for our images.

\newcommand{\pathSigma}{./figures/Fig16JPG/}
\newcommand{\imSigma}[1]{\includegraphics[width=1.6cm]{\pathSigma #1.jpg}}
\begin{figure}
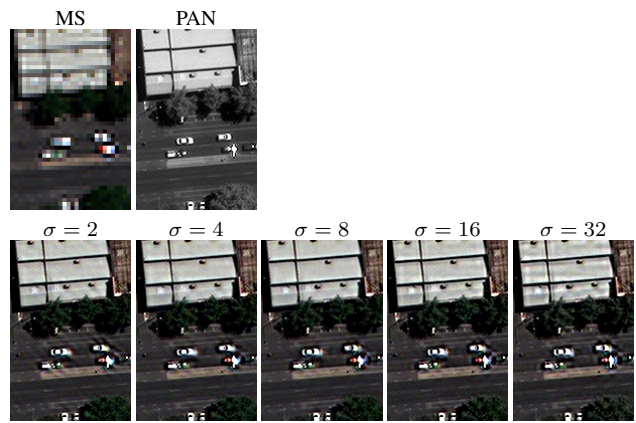

\centering
\tiny
\setlength{\tabcolsep}{1pt}
\begin{tabular}{ccccc}
       \footnotesize MS  &        \footnotesize PAN  &                          &                           &                           \\ \vspace{1mm}
            \imSigma{MS} &             \imSigma{PAN} &                          &                           &                           \\
\footnotesize $\sigma=2$ &  \footnotesize $\sigma=4$ & \footnotesize $\sigma=8$ & \footnotesize $\sigma=16$ & \footnotesize $\sigma=32$ \\ \vspace{1mm}
       \imSigma{scale_2} &         \imSigma{scale_4} &        \imSigma{scale_8} &        \imSigma{scale_16} &        \imSigma{scale_32} \\
\end{tabular}
\caption{Impact of patch size ($\sigma$) on pansharpening quality.}
\label{fig:sigma}
\end{figure}

\subsubsection{Setting $\beta$}
\begin{figure}
\centering
\includegraphics[width = 0.92\columnwidth]{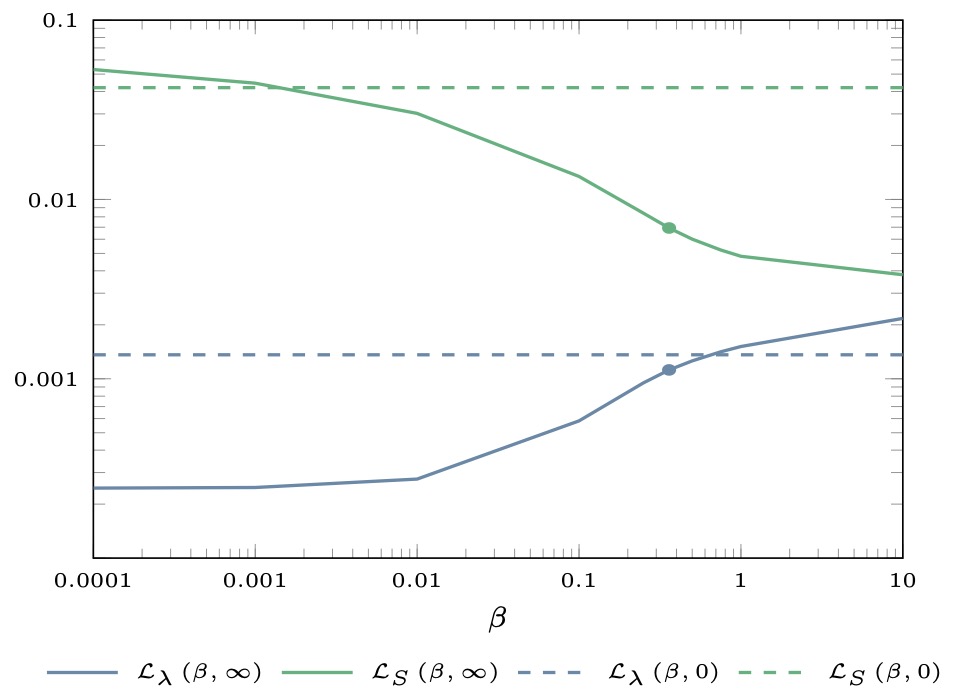}
\caption{Spatial and spectral losses as a function of $\beta$.}
\label{fig:losses_vs_beta}
\end{figure}

the parameter $\beta$ balances the relative importance of the spatial and spectral loss terms.
When $\beta=0$, only the spectral loss is taken into account, which negatively affects the spatial quality, and the opposite happens when $\beta \to \infty$.
To quantify this behavior,
Fig.\ref{fig:losses_vs_beta} reports the values of the spatial and spectral loss terms observed for A-PNN-TA-FR when $\beta$ grows from 0.0001 to 10.
There is a large range of values where both losses (solid lines) decrease with respect to the case without adaptation (dashed lines).
So, to gain a better insight, we resort again to visual inspection for a sample test image.
In Fig.\ref{fig:beta} we show the original MS (enlarged) and PAN, in the first row,
the pansharpened outputs for various values of $\beta$, in the second row,
and the difference between the former and an interpolated version of the MS, in the third row.
For $\beta=0.01$ and even 0.1, the output images appear blurred, with an insufficient spatial quality.
For $\beta=10$, instead, and to a lesser extent also for $\beta=1$, there are color distortions on the vegetation and other details,
especially visible in the difference images.
A good compromise is obtained with values between 0.1 and 1,
and in fact we selected eventually $\beta=0.25$ for GeoEye and $\beta=0.36$ for WorldView.

\newcommand{\pathBeta}{./figures/Fig18JPG/}
\newcommand{\imBeta}[1]{\includegraphics[width=1.65cm]{\pathBeta #1.jpg}}
\begin{figure}
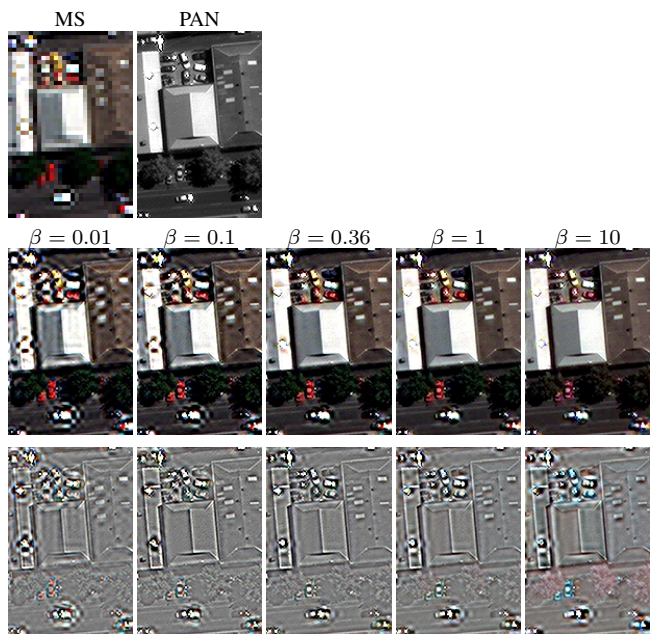

\centering
\tiny
\setlength{\tabcolsep}{1pt}
\begin{tabular}{ccccc}
         \footnotesize MS  &        \footnotesize PAN  &                            &                         &                          \\ \vspace{1mm}
               \imBeta{MS} &              \imBeta{PAN} &                            &                         &                          \\
\footnotesize $\beta=0.01$ & \footnotesize $\beta=0.1$ & \footnotesize $\beta=0.36$ & \footnotesize $\beta=1$ & \footnotesize $\beta=10$ \\ \vspace{1mm}
        \imBeta{beta_0.01} &         \imBeta{beta_0.1} &         \imBeta{beta_0.36} &         \imBeta{beta_1} &         \imBeta{beta_10} \\
      \imBeta{D_beta_0.01} &       \imBeta{D_beta_0.1} &       \imBeta{D_beta_0.36} &       \imBeta{D_beta_1} &       \imBeta{D_beta_10} \\
\end{tabular}
\caption{Impact of loss balance ($\beta$) on pansharpening quality.}
\label{fig:beta}
\end{figure}

\subsubsection{Testing an alternative loss}
our system has been conceived based on a clear rationale, discussed in section 3, and our training loss was designed to fulfil it.
Nonetheless, one can legitimately wonder what happens if different losses are used in the same framework.
So we fine-tuned the A-PNN-TA-FR model replacing our loss with a very different one, recently proposed \cite{Luo2020} for high-resolution training.

This latter, call it $\LL'$ comprises four terms
\begin{equation}
    \LL' = \LL'_{\lambda} + \LL'_{S} + \LL'_{\rm QNR} + \lambda \parallel \Theta \parallel^2_2
\label{eq:tot_competitor}
\end{equation}
The first two aim at improving spectral and spatial quality by minimizing combinations of MSE and structural similarity indexes in the upscaled multispectral and panchromatic domains.
Instead, the third term directly targets the QNR \cite{Vivone2015} a well-known full resolution quality measure,
while the last one, the weighted norm of the parameters, serves only for regularization.
The reader is referred to \cite{Luo2020} for all details.

\newcommand{\pathC}{./figures/Fig19JPG/}
\newcommand{\imComp}[1]{\includegraphics[width=2.0cm]{\pathC #1.jpg}}
\begin{figure}
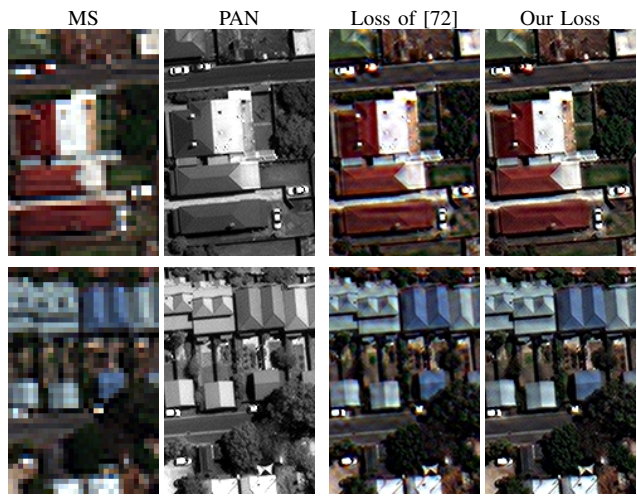

\centering
\tiny
\setlength{\tabcolsep}{1pt}
\begin{tabular}{cc@{\rule{2mm}{0mm}}ccc}
\footnotesize MS & \footnotesize PAN & \footnotesize Loss of \cite{Luo2020} & \footnotesize Our Loss \\ \vspace{1mm}
\imComp{MS_1} & \imComp{PAN_1} & \imComp{C1_1} & \imComp{A-PNN-FT-Z_1} \\ \vspace{1mm}
\imComp{MS_3} & \imComp{PAN_3} & \imComp{C1_3} & \imComp{A-PNN-FT-Z_3}
\end{tabular}
\caption{Comparing the loss of \cite{Luo2020} with the proposed loss.}
\label{fig:comparing_losses}
\end{figure}

Sample pansharpened crops are shown in Fig.\ref{fig:comparing_losses}, next to the original MS and PAN, for two samples for the WorldView3 Adelaide image.
The spectral fidelity is quite good in both cases, although slightly better indexes are obtained with our loss,
$\DL=0.03$ as opposed to 0.05.
When considering spatial fidelity, however, an obvious performance gap appears.
Images pansharpened with our loss are much sharper,
fine textures and small details are much better preserved, as obvious from the comparison with the PAN,
and no spurious pattern is generated in the process.

\subsection{Strengths and weaknesses of the proposed framework}
Results of the previous Subsections make clear what the main strengths of the proposed framework are.
By using the original PAN-MS pairs to train a deep learning model, we make sure that the most informative data are taken into account
and lay the basis for obtaining high spectral and spatial fidelity in pansharpening.
Results obtained with A-PNN, PanNet, and DRPNN are just examples of the potential of this approach.
On the down side, working at high resolution incurs costs.
Using the original data, without subsampling, causes pre-training to become much slower and memory intensive,
a nuisance, but not a major problem, considering that pre-training takes place off-line.
On the other hand, target adaptation is important to ensure the best performance, and this process takes place on-line.
For Z-PNN and 100 iterations it requires about three minutes, as shown in Tab.\ref{tab:costs}.
Depending on application and mode of use, this may be overly annoying.
With the following experiment, however, we show that this cost may be significantly reduced.

Fig.\ref{fig:zoom} shows, on the left column, the original PAN-MS pair for a 128$\times$128-pixel WV3 crop.
In the middle column, we see the output of A-PNN-TA on the top and Z-PNN on the bottom, both adapted on the 2048$\times$2048-pixel target image including our crop,
displaying the by-now usual quality gap.
Our focus, though, is on the right column.
Here, adaptation is carried out {\em only} on the very same 128$\times$128-pixel target crop, not the whole image, hence, using much less data and computing time.
While the quality of the A-PNN-TA image further degrades, likely for the lack of sufficient data,
this is not the case for the Z-PNN image, which is almost indistinguishable from the previous case.
As this behavior is observed consistently in our experiments, we conclude that Z-PNN can be safely fine tuned on the very same scene of interest, even very small, providing stable and high-quality results.
Needless to say, this comes at a fraction of the original computational cost, just about 1 second in our example.
Therefore, one can use Z-PNN in this modality to ``zoom'' on the details of interest,
each time upgrading the original Z-PNN output (already good) in a matter of seconds.

Another critical point regards the different speed of adaptation of the spectral and spatial loss terms (see Fig.~\ref{fig:loss}).
Since the latter improves much faster than the former,
Z-PNN and A-PNN-TA-FR turn out to have a very similar spatial score ($\DR$) but, in some cases, a non negligible gap in terms of spectral score ($\DL$, R-SAM).
This may call for a longer adaptation phase in the presence of very strict spectral accuracy requirements.

A valuable strength of the proposed framework is the automatic co-registration of pansharpened spectral bands.
To better appreciate this feature, in Fig.\ref{fig:coregistration} we show, for another WorldView-3 150$\times$150-pixel crop,
the input PAN-MS pair and the output images generated by Z-PNN and some reference methods where the co-registration problem is not addressed.
This time, however, we use an unusual red-yellow-blue false-color representation.
In fact, while the red, green, and blue bands are usually well aligned,
other bands, such as the yellow one, may be slightly shifted, due to the imaging system that acquires subsets of bands in slightly different time intervals.
As expected, severe color distortions are visible in all the output images except for Z-PNN,
where spectral fidelity remains high also near sharp boundaries.

\newcommand{\pathZoom}{./figures/Fig20JPG/}
\newcommand{\imZoom}[1]{\includegraphics[width=2.8cm]{\pathZoom #1.jpg}}
\begin{figure}
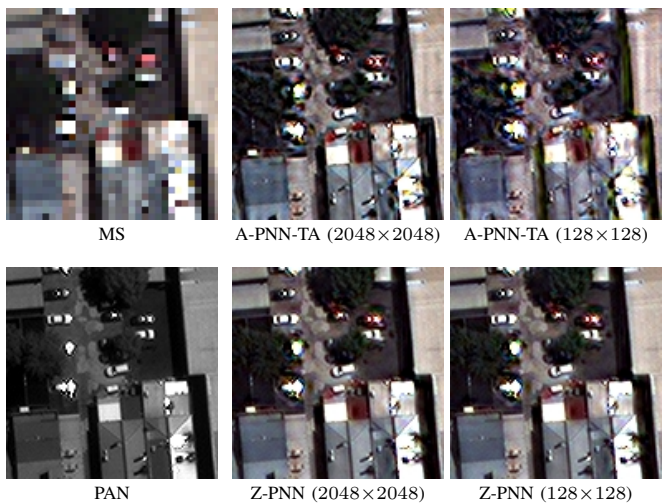

\setlength{\tabcolsep}{1pt}
\centering
\scriptsize
\begin{tabular}{c@{\rule{2mm}{0mm}}c@{\rule{1mm}{0mm}}c}
 \imZoom{MS} &           \imZoom{a_pnn_2048} & \imZoom{a_pnn_128}           \\
          MS & A-PNN-TA ($2048{\times}2048$) & A-PNN-TA ($128{\times}128$)  \\ \\
\imZoom{PAN} &           \imZoom{z_pnn_2048} & \imZoom{z_pnn_128}           \\
         PAN &    Z-PNN ($2048{\times}2048$) & Z-PNN ($128{\times}128$)     \\ [2mm]
\end{tabular}
\caption{
For Z-PNN, the quality of fine tuning does not appreciably depend on the size of the target scene, 2048$\times$2048 (middle) or 128$\times$128 (right).
Therefore, it can be used to ``zoom'' in real time on any detail of interest.}
\label{fig:zoom}
\end{figure}

\newcommand{\pathCor}{./figures/Fig21JPG/}
\newcommand{\imCor}[1]{\includegraphics[width=2.8cm]{\pathCor #1.jpg}}
\begin{figure}
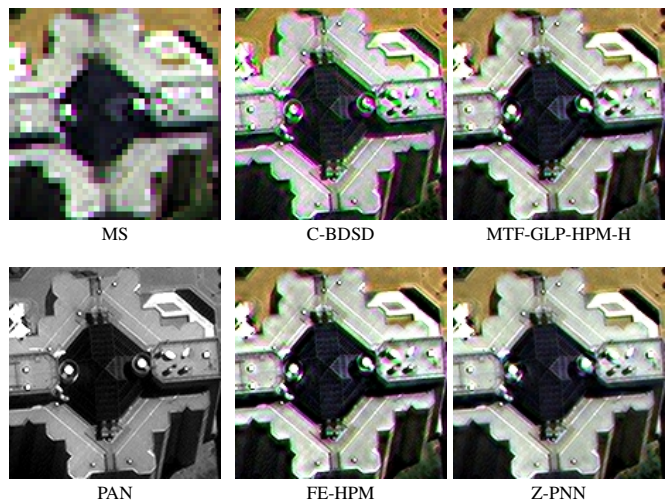

\setlength{\tabcolsep}{1pt}
\centering
\scriptsize
\begin{tabular}{c@{\rule{2mm}{0mm}}c@{\rule{1mm}{0mm}}c}
 \imCor{MS} & \imCor{C-BDSD} &      \imCor{MTF-GLP-HPM-H} \\
        MS  &        C-BDSD  &             MTF-GLP-HPM-H  \\ \\
\imCor{PAN} & \imCor{FE-HPM} & \imCor{Z-PNN-Coregistered} \\
       PAN  &        FE-HPM  &        Z-PNN               \\ [2mm]
\end{tabular}
\caption{In the red-yellow-blue display, the effects of MS bands misalignment is highlighted.
Spourious green or magenta lines appear along object borders in all pansharpened images except Z-PNN's, where this issue  is automatically addresses.
}
\label{fig:coregistration}
\end{figure}

To conclude this Subsection, in Fig.\ref{fig:WV3FRcrops_veg} we show the output of {\em all} reference methods for a single small vegetation crop.
Vegetation is extremely common in multi-resolution imagery, but its correct pansharpening is often prohibitive due to the presence of fine textures at multiple scales.
This is confirmed by the results in the figure.
Apart from some methods that present a clear failure ({\it e.g.} DRPNN*) many more provide disappointing results,
with large chromatic aberrations and/or a significant loss of detail.
In general, MRA methods perform quite well on this image, much better than CS and VO.
Also ML methods trained at low resolution are among the worst in this task.
Instead, thanks to the spatial loss based on local correlation,
A-PNN-TA-FR and Z-PNN, trained with our high resolution framework, show again a very good performance, preserving faithfully even the most subtle vegetation textures.

\newcommand{\pathVEG}{./figures/Fig22JPG/}
\newcommand{\imVEG}[1]{\includegraphics[width=1.75cm]{\pathVEG #1.jpg}}
\begin{figure*}
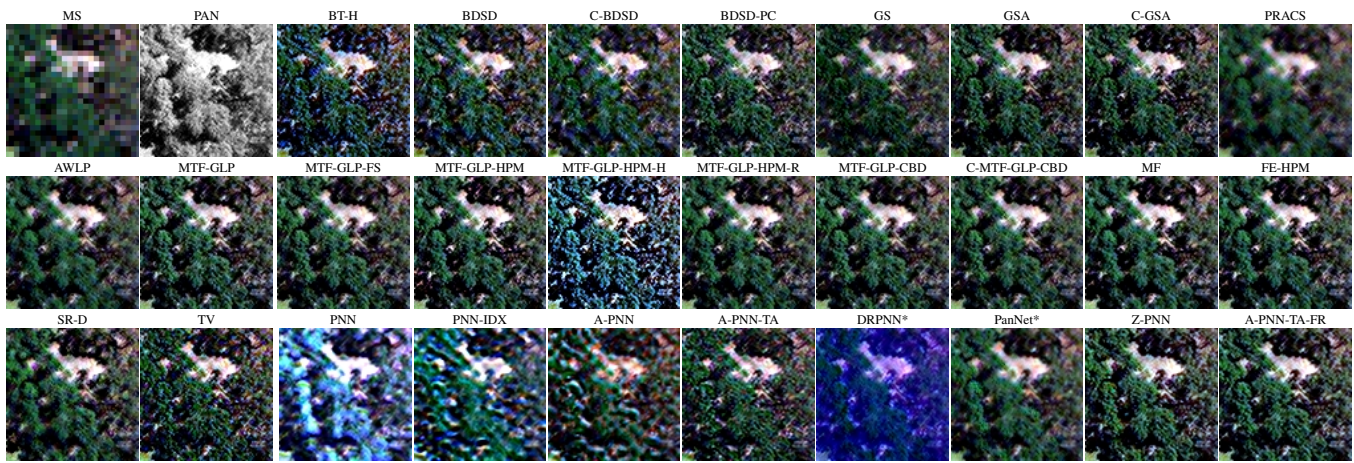

\centering
\tiny
\setlength{\tabcolsep}{0.5pt}
\begin{tabular}{cccccccccc}
          MS &             PAN &               BT-H &                BDSD &                C-BDSD &               BDSD-PC &                  GS &                   GSA &         C-GSA &                PRACS \\
  \imVEG{MS} &     \imVEG{PAN} &       \imVEG{BT-H} &        \imVEG{BDSD} &        \imVEG{C-BDSD} &       \imVEG{BDSD-PC} &          \imVEG{GS} &           \imVEG{GSA} & \imVEG{C-GSA} &        \imVEG{PRACS} \\
        AWLP &         MTF-GLP &         MTF-GLP-FS &         MTF-GLP-HPM &         MTF-GLP-HPM-H &         MTF-GLP-HPM-R &         MTF-GLP-CBD &         C-MTF-GLP-CBD &            MF &               FE-HPM \\
\imVEG{AWLP} & \imVEG{MTF-GLP} & \imVEG{MTF-GLP-FS} & \imVEG{MTF-GLP-HPM} & \imVEG{MTF-GLP-HPM-H} & \imVEG{MTF-GLP-HPM-R} & \imVEG{MTF-GLP-CBD} & \imVEG{C-MTF-GLP-CBD} &    \imVEG{MF} &       \imVEG{FE-HPM} \\
        SR-D &              TV &                PNN &             PNN-IDX &                 A-PNN &              A-PNN-TA &        DRPNN* &        PanNet* &          Z-PNN &        A-PNN-TA-FR \\
\imVEG{SR-D} &      \imVEG{TV} & \      \imVEG{PNN} &     \imVEG{PNN-IDX} &         \imVEG{A-PNN} &      \imVEG{A-PNN-FT} &           \imVEG{DRPNN} & \imVEG{PanNet} &  \imVEG{Z-PNN} & \imVEG{A-PNN-Zoom-FT}
\end{tabular}
\caption{Results of all methods on a small WV3 vegetation crop.
Most methods, including ML methods trained at low resolution, show chromatic aberrations and resolution loss.
ML methods trained at high resolution ensure high spatial and spectral fidelity.}
\label{fig:WV3FRcrops_veg}
\end{figure*}

\subsection{Implementation details}

All experiments were run on a server equipped with Nvidia Quadro P6000 GPU with 24GB memory, and all networks were implemented in PyTorch.
Some of the tested CNN models, {\em i.e.}, Z-PNN, PanNet*, DRPNN*, needed a pretraining phase.
For Z-PNN, as stated in Sec.~\ref{sec:zpnn},
the model weights have been produced as refinement of the original A-PNN model parameters \cite{Scarpa2018a},
using a dedicated training image for each sensor, as indicated in Tab.~\ref{tab:datasets}.
The whole image is used as a one-sample batch,
running 2000 iterations that involve all layers, with a learning rate of $10^{-5}$ on WorldView-2/3 and of $5\cdot10^{-5}$ on GeoEye-1,
and using the Adam optimizer with $\beta_1=0.9$ and $\beta_2=0.99$.

The models for PanNet* and DRPNN*, instead,
have been reimplemented in PyTorch and trained from scratch on our training datasets for all three sensors,
using the same hyperparameters (learning rate, optimizer, loss, epochs, etc.) of the original versions.
For these models, however, since the training occurs in the reduced resolution domain,
we used a 4$\times$4 times larger tile, hence 8192$\times$8192 pixels, to compensate for data volume reduction.

\section{Conclusion}
We have proposed a framework for full-resolution training of pansharpening models, with the aim of exploiting all the information carried by the original data, with no resolution downgrading.
Lacking a ground truth, we defined a suitable compound loss, with two components accounting separately for spectral and spatial fidelity.
We used the proposed framework to train several state-of-the-art pansharpening models.
Experimental results are extremely encouraging.
Besides numerical indicators, visual inspection confirms that the quality of the pansharpened images is largely improved thanks to high-resolution training.
Beyond the framework itself and the trained pansharpening methods, though, the main contribution of this work is to prove the potential of this training approach.
Many improvements are certainly possible, and we hope to stimulate research on this topic.
We are currently working on a refined spatial loss component.

\balance

\bibliographystyle{IEEEtran}
{\footnotesize \bibliography{refs}}

\end{document}